\documentclass[letterpaper, 10 pt, conference]{ieeeconf}  

\IEEEoverridecommandlockouts                              

\usepackage{times}
\usepackage{epsfig}
\usepackage{graphicx}
\usepackage{amsmath}
\usepackage{amssymb}
\usepackage{subcaption}
\usepackage{url}
\usepackage{booktabs}
\usepackage{multirow}
\usepackage{colortbl}
\usepackage[dvipsnames]{xcolor}
\usepackage{verbatim}
\usepackage{color}
\usepackage{verbatim}

\usepackage[breaklinks=true,bookmarks=false,colorlinks,urlcolor=black, backref=page]{hyperref}
\newcommand\email[1]{\href{mailto:#1}{\nolinkurl{#1}}}

\usepackage[normalem]{ulem}
\definecolor{mygray}{gray}{0.8}

\def\eg{\emph{e.g.}} 
\def\ie{\emph{i.e.}}

\def\etal{\emph{et al.}}

\definecolor{myYellow}{RGB}{230, 180, 0}

\title{\LARGE \bf
UniFuse: Unidirectional Fusion for \boldmath$360^{\circ}$ Panorama Depth Estimation}

\author{Hualie Jiang$^{1}$, Zhe Sheng$^{2}$, Siyu Zhu$^{2}$, Zilong Dong$^{2}$ and Rui Huang$^{1}$ 
\thanks{$^{1}$Shenzhen Institute of Artificial Intelligence and Robotics for Society, The Chinese University of Hong Kong, Shenzhen.}
\thanks{$^{2}$Alibaba Cloud A.I. Lab. This work was mainly done when Hualie Jiang interned at Alibaba Cloud A.I. Lab.}
}

\begin{document}

\maketitle
\thispagestyle{empty}
\pagestyle{empty}

\begin{abstract}
Learning depth from spherical panoramas is becoming a popular research topic because a panorama has a full field-of-view of the environment and provides a relatively complete description of a scene. However, applying well-studied CNNs for perspective images to the standard representation of spherical panoramas, \ie, the equirectangular projection, is suboptimal, as it becomes distorted towards the poles. Another representation is the cubemap projection, which is distortion-free but discontinued on edges and limited in the field-of-view. This paper introduces a new framework to fuse features from the two projections, unidirectionally feeding the cubemap features to the equirectangular features only at the decoding stage. Unlike the recent bidirectional fusion approach operating at both the encoding and decoding stages, our fusion scheme is much more efficient. Besides, we also designed a more effective fusion module for our fusion scheme. Experiments verify the effectiveness of our proposed fusion strategy and module, and our model achieves state-of-the-art performance on four popular datasets. Additional experiments show that our model also has the advantages of model complexity and generalization capability. The code is available at~\url{https://github.com/alibaba/UniFuse-Unidirectional-Fusion}. 
\end{abstract}


\section{INTRODUCTION}

Depth estimation is a fundamental step in 3D reconstruction, having many applications, such as robot navigation and virtual/augmented reality.  
A spherical (or $360^{\circ}$, omnidirectional) panoramic image has a full field-of-view of the environment, thus has the potential to produce a more accurate, complete, and scale-consistent reconstruction of scenes.
This paper presents our work on better predicting depth from a single spherical panoramic image. 

The $360^{\circ}$ panorama is usually represented as the equirectangular projection (ERP) or cubemap projection (CMP)~\cite{skupin2017standardization}. Both of them are different from the perspective image and have their respective advantages and disadvantages. 
EPR provides a complete view of the scene but contains distortion that becomes severer towards the poles. In contrast, CMP is distortion-free but discontinued on face sides and limited in the field-of-view.
Applying deep CNNs to panoramic images for accurate depth estimation is thus challenging. 

Recently, BiFuse~\cite{wang2020bifuse} combines the two above projections for depth estimation, which builds bidirectional fusion of the two branches at both the encoding and decoding stages and finally uses a refinement network to fuse the estimated depth maps from both branches. To alleviate the discontinuity of CMP, BiFuse also adopts the spherical padding among cube faces. However, with too many modules added, BiFuse becomes over-complicated, as discussed in detail in Sec.~\ref{sec:ca}. We argue that feeding the ERP features to the CMP branch is unnecessary, as the ultimate goal is to output an equirectangular depth map.
Optimizing the cube map depth may cause the training to lose focus on the equirectangular depth. Furthermore, performing the fusion at the encoding stage may disturb the learning of the encoder, as it is usually initialized with ImageNet~\cite{deng2009imagenet} pretrained parameters.

\begin{figure*}[t]
\vspace{6pt}
\begin{center}
\includegraphics[width=0.86\linewidth]{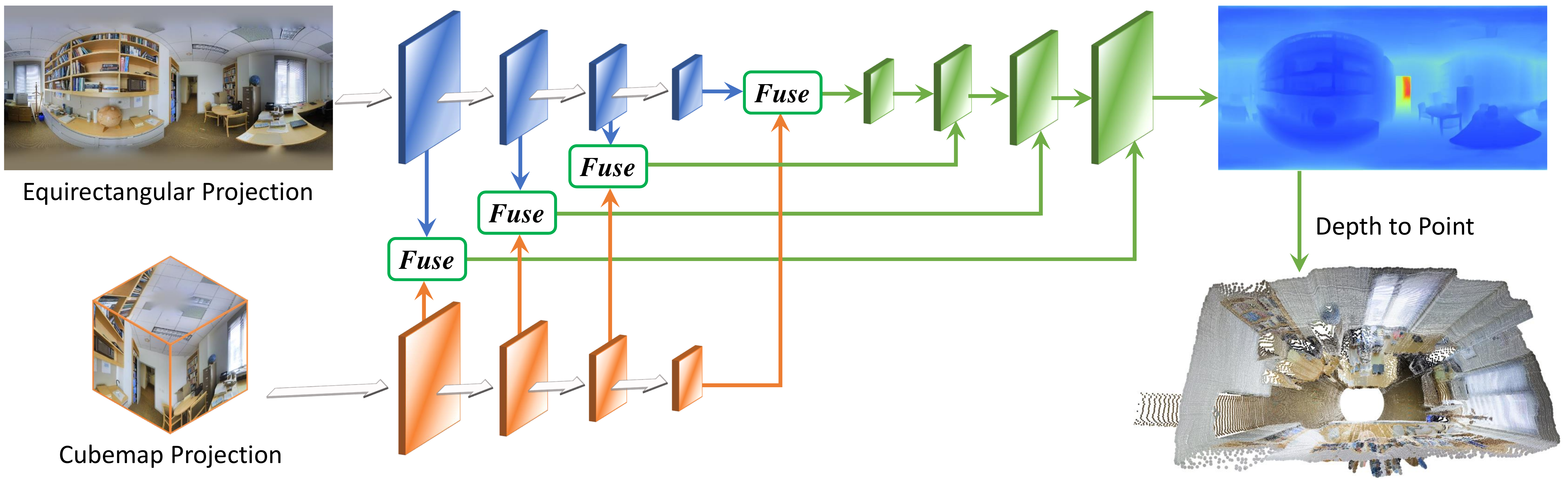}
\end{center}
\caption{\textbf{Our Proposed Unidirectional Fusion Framework.} }
\label{fig:framework}
\end{figure*}

To address the above limitations, we propose a new fusion framework, which unidirectionally feeds the features extracted from CMP to the ERP branch only at the decoding stage to better support the ERP  prediction, as shown in Fig.~\ref{fig:framework}. The fusion scheme uses the simple U-Net~\cite{ronneberger2015u} and performs the fusion at the skip connections so that the fusion has minimum coupling with the backbones.
Besides, we design a fusion module for our fusion framework, aiming at using Cubemap to Enhance the Equirectangular features (noted as CEE). We first adopt a residual modulation of the cubemap features to mitigate its discontinuity. Because the concatenation of modulated cubemap features and equirectangular features doubles the feature map channels, to better model the channel-wise importance, we introduce the Squeeze-and-Excitation (SE) \cite{hu2018squeeze} block to the CEE module. Our CEE module works better for our fusion scheme than the simple concatenation or the Bi-Projection~\cite{wang2020bifuse} module does.

Our contributions are summarized as follows: 
(1) we propose a new fusion framework of equirectangular and cubemap features for single spherical panorama depth estimation, 
(2) we design a better fusion module for our unidirectional fusion framework than existing modules, 
and (3) we perform experiments to show our approach's effectiveness, and the final model establishes the state-of-the-art performance and has advantages on the complexity and generalization ability.

\section{RELATED WORK}
\label{sec:rw}

Make3D~\cite{saxena2009make3d} is a seminal work on a single perspective image depth estimation, which uses the traditional graphical model.  
With the development of deep learning, convolutional neural networks were applied to this task \cite{eigen2014depth, liu2015learning, laina2016deeper, jiang2019high, fu2018deep, li2018monocular, jiang2019hierarchical}. This task is either treated as a dense regression problem~\cite{eigen2014depth, liu2015learning, laina2016deeper, jiang2019high} or a classification problem~\cite{fu2018deep, li2018monocular, jiang2019hierarchical} by discretizing the depth. The experiments are usually performed on datasets with ground truth depth obtained with physical sensors. 
To avoid the direct usage of ground truth depth, some work tried to utilize other data source for training, \eg, stereo images~\cite{garg2016unsupervised, godard2017unsupervised, godard2019digging}, monocular videos~\cite{zhou2017unsupervised, wang2019unsupervised, godard2019digging, jiang2020dipe}, where the training objective is to minimize the between-view reconstruction error. However, the performance of these unsupervised approaches is inferior to the supervised ones.

The spherical panorama has a full field-of-view of a scene, which can extract more accurate and scale-consistent depth than the perspective image. Zioulis~\etal~\cite{zioulis2018omnidepth} first performed depth estimation on panoramas and proposed to replace the first two layers of the network with a set of rectangle filters~\cite{su2017learning} to handle distortion.
They constructed the 3D60 dataset rendered from several datasets. But this dataset is relatively easy due to a problem of rendering, as pointed out in Sec.~\ref{sec:pc}. 
Later, they constructed both vertical and horizontal stereo panoramas to perform unsupervised $360^{\circ}$ depth learning~\cite{zioulis2019spherical}. Similarly, Wang~\etal~\cite{wang2018self} composed a purely virtual panorama dataset PonoSUNCG with panorama video frames to perform unsupervised depth learning like SfMLearner~\cite{zhou2017unsupervised}. 
Pano Popups~\cite{eder2019pano} jointly learns depth with surface normals and boundaries to improve depth estimation. More recently, ODE-CNN~\cite{cheng2020ode} reduces the $360^{\circ}$ depth estimation problem as an extension problem from the front face depth. Both their experiments are still performed on virtual datasets only.

However, virtual datasets tend to be too easy, and the models trained on them are probably hard to transfer to real applications. 
Tateno~\etal~\cite{tateno2018distortion} first experimented on the real dataset Stanford2D3D~\cite{armeni2017joint}. They proposed to train on the common sub perspective views and then transfer to the full panorama images by applying a distortion-aware deformation on the convolutional kernels of the trained model. Though such a method has the potential to utilize more available RGBD datasets to learn a panorama depth estimation model, it fails to take advantage of the large receptive field-of-view of the panorama. More recent work~\cite{jin2020geometric, wang2020bifuse} tends to build a complex model on this task. Jin~\etal~\cite{jin2020geometric} leveraged the layout elements as both the prior and regularizer for depth estimation, resulting in a model with three encoders and seven decoders. In comparison, our UniFuse contains only two encoders and one decoder. Wang~\etal~\cite{wang2020bifuse} first proposed to utilize both EPR and CMP for $360^{\circ}$ depth estimation. Their model BiFuse is composed of two networks, \ie, the equirectangular and cubemap branches, between which are bidirectional fusion modules. There is also a refinement network to refine the predicted depth maps of the two branches. 
However, its complex structures may hinder the concentration on the learning of equirectangular features, which is critical to the final equirectangular depth map. 
In contrast, both our unidirectional fusion framework and the CEE fusion module are designed to use the cubemap to enhance the equirectangular feature learning. 

{There are some existing elaborate convolutions for handling distortion of EPR and special padding techniques for discontinuity of both EPR and CMP. 
The convolutions include the Spherical Convolution (SC)~\cite{su2017learning} which is a set of rectangle filters and is used at the front layers of the network, and the Distortion-aware Convolution (DaC)~\cite{tateno2018distortion, coors2018spherenet, fernandez2020corners} which samples the features for convolution from a regular grid on the tangent plane instead of EPR. The padding methods include the Circular Padding (CirP)~\cite{wang2018omnidirectional} for EPR and Cube Padding (CuP)~\cite{cheng2018cube} and Spherical Padding (SP)~\cite{wang2020bifuse} for CMP. Readers could refer to their original papers for more technical details. We do not adopt these special convolutions and paddings in our models. Experiments in Sec.~\ref{sec:ca} show that using these methods in UniFuse does not improve the performance but usually adds the complexity.}

\section{METHODOLOGY}
\subsection{Preliminaries}
In this section, we introduce the two common projections for the spherical image, i.e., the equirectangular projection and cubemap projection, and their mutual conversion. 

\textbf{Equirectangular Projection} is the representation of a spherical surface by uniformly sampling it in longitudinal and latitudinal angles. The sampling grid is rectangular, in which the width is twice of the height. Suppose that the longitude and latitude are $\phi$ and $\theta$ respectively, and we have $(\phi,  \theta) \in [0, 2\pi] \times [0, \pi]$. The angular position $(\phi,  \theta)$ can be converted into the coordinate $P_s=(p_s^x, p_s^y, p_s^z)$ in the standard spherical surface with radius $r$ by, 

\begin{equation}
\begin{array}{rl}
	&p_s^x= r \sin(\phi)\cos(\theta), \\
	&p_s^y= r \sin(\theta), \\
	&p_s^z= r \cos(\phi)\cos(\theta).
\end{array}
\label{eq:equi}
\end{equation}

\begin{figure*}[t]
\vspace{5pt}
\begin{center}
\includegraphics[width=0.96\linewidth]{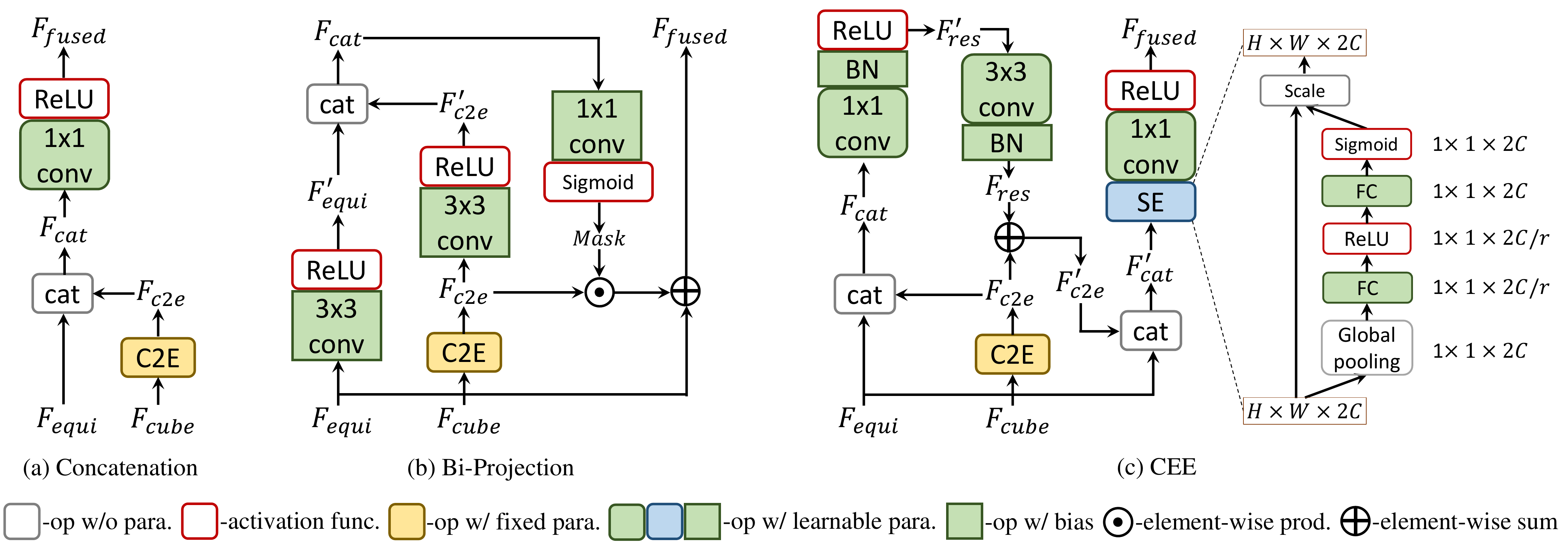}
\end{center}
\caption{\textbf{The Fusion Modules.} }
\label{fig:modules}
\end{figure*}

\textbf{Cubemap Projection} is the projection of a spherical surface to the $6$ faces of its inscribed cube. The $6$ faces are specific perspective images, whose size is $r \times r$ and focal length $r/2$. The $6$ faces can be denoted as $f_i, i \in {B, D, F, L, R, U}$, corresponding to the looking directions, $-z$(back), $-y$(down), $z$(front), $x$(left), $-x$(right) and $y$(up). The front face has the identical coordinate system with the spherical surface, while others have either $90^{\circ}$ or $180^{\circ}$ rotations around one axis. Let us denote the rotation matrix from the system of the spherical surface to one of the i-\textit{th} face as $R_{f_i}$. Then we can project the pixel $P_c=(p_c^x, p_c^y, p_c^z)$ in $f_i$ by, 
\begin{equation}
P_s = s \cdot R_{f_i}  P_c, 
\label{eq:cube}
\end{equation}
where, $p_c^x, p_c^y \in [0, r]$, $p_c^z=r/2$, and the factor $s=r/|p_c|$.

\textbf{C2E} is the reprojection of the contents (raw RGB values or features) in cubemaps to the equirectangular grid. \textbf{C2E} is usually performed as an inverse wrapping, where we have to compute the corresponding point with the angular position $(\phi, \theta)$. Specifically, we first use Equ.~(\ref{eq:equi}) to reproject the angular position to the spherical surface. Then we determine the projected face by finding the minimum angular distance between it and the looking directions of cubemaps. Finally, we compute the corresponding position in the cube face by using the inverse process of Equ.~(\ref{eq:cube}).

\subsection{The Unidirectional Fusion Network}
Our unidirectional fusion network of ERP and CMP is illustrated in Fig~\ref{fig:framework}.
The reason to perform fusion in a unidirectional manner is that
the ultimate goal of $360^{\circ}$ depth estimation is to produce an equirectangular depth map, and to feed distortion-free cubemap features to the full-view equirectangular features as a supporting component is a natural choice. We do not perform fusion in a reverse direction, as CMP is limited in the field-of-view and the spherical padding~\cite{wang2020bifuse} for the discontinuity of CMP is time-consuming. 
Additionally, the decoder for the CMP branch to predict cube depth maps increases the complexity, and optimizing the cube depth maps would distract the learning of equirectangular depth. Therefore, we do not adopt a decoder for the cubemaps, and the network contains only two encoders and one decoder. 
To avoid disturbing the learning of the backbone, we choose to perform fusion only at the decoding stage, and it is better to fuse the well-encoded features. To this end, we adopt a U-Net~\cite{ronneberger2015u} as a baseline network and perform fusion within skip-connections.

\subsection{The Fusion Modules}
In this section, we introduce our proposed fusion module for the UniFuse framework, as well as two baseline fusion methods, concatenation, and the Bi-Projection \cite{wang2020bifuse}, as illustrated in Fig.~\ref{fig:modules}. The dimensions of common features of these 3 modules are,
\begin{itemize}
\item $F_{equi}/F'_{equi}$, $F_{c2e}/F'_{c2e}$, $F_{res}/F'_{res}$, $F_{fused}$: $H \times W \times C$
\item $F_{cube}$: $ 6 \times H/2 \times H/2 \times C$
\item $F_{cat}$/$F'_{cat}$: $H \times W \times 2C$
\end{itemize}
where $H$, $W$ and $C$ are height, width, and channel of the equirectangular features, and cubemap features have the same channels, but the face size is just $H/2$.

\textbf{Concatenation} module first casts the CMP features to EPR by \textbf{C2E}, then concatenates them with $F_{equi}$ and finally uses a $1 \times 1$ conv. module to reduce the channel from $2C$ to $C$. The number of parameters is $2C^2$. 

\textbf{Bi-Projection} \cite{wang2020bifuse} aims at generating a masked feature map from one branch and add it to another one. In (b) of Fig.~\ref{fig:modules}, we omit the \textbf{E2C} path, as it is not necessary in our unidirectional fusion framework. The $Mask$ is $H \times W \times 1$. To generate the mask, the Bi-Projection first uses two $3 \times 3$ conv. modules to encode $F_{equi}$ and $F_{c2e}$ as  $F'_{equi}$ and $F'_{c2e}$, then reduce the concatenated feature map's channel to $1$, and finally apply a Sigmoid function to scale the $Mask$ between $0$ and $1$. Such masked modification in Bi-Projection may be useful in gradually improving the feature learning in two branches in BiFuse, but seems not effective enough for our unidirectional fusion framework at the decoding stage. The number of parameters is $18C^2 + 4C + 1$. 

\begin{figure}[t]
\vspace{5pt}
\begin{center}
\includegraphics[width=\columnwidth]{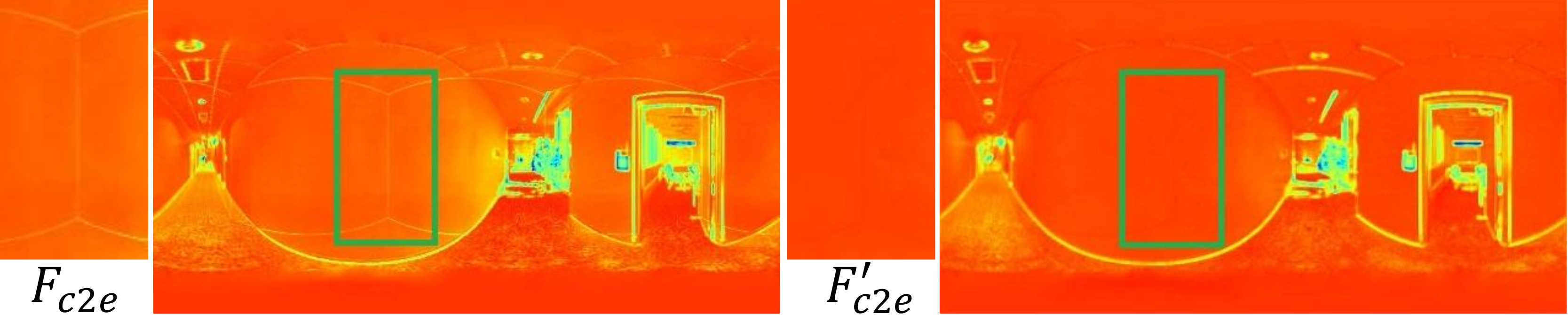}
\end{center}
\caption{\textbf{The Visualization of} $F_{c2e}$ and $F'_{c2e}$.}
\label{fig:feat}
\end{figure}

\textbf{CEE} is a more elaborate concatenation that better facilitates the fusion process. It aims at using the distortion-free cubmap to enhance the equirectangular features. Because the cubemap features are probably inconsistent in cubemap boundaries, we first generate a residual feature map $F_{res}$ to be added to $F_{c2e}$ to reduce such an effect. To generate $F_{res}$, we design a residual block inspired by ResNet~\cite{he2016deep} to the concatenation of $F_{equi}$ and $F_{c2e}$. The residual block contains a $1 \times 1$ conv. module to squeeze the channels and a $3 \times 3$ conv. module to generate the residual feature map. 
{Fig.~\ref{fig:feat} shows the feature map of $F_{c2e}$ and $F'_{c2e}$ at the $1/2$ resolution stage. Cracks appear between cube faces in $F_{c2e}$ but they disappear in $F'_{c2e}$, which indicates that the residual modulation has filled them. An intuitive explanation is that the continuous $F_{equi}$ helps the residual block localize inconsistent boundaries of $F_{c2e}$ and the supervision from continuous ground truth helps learn to generate values for filling boundaries.}
The remained part is similar to \textbf{Concatenation}. As we have dual channels of features from both branches, before the final $1 \times 1$ conv. module, we add a Squeeze-and-Excitation (SE) \cite{hu2018squeeze} block, which can adaptively recalibrates channel-wise feature responses, and thus masking the fusion better. We set r in the SE block as 16, so the total number of parameters is $13.5C^2+ 4C$. Therefore the number of parameters in our CEE is considerably smaller (75\%) than that of the Bi-Projection.


\section{EXPERIMENTS}

\subsection{Experimental Settings}

\subsubsection{Datasets}
Our experiments are conducted on four datasets, Matterport3D~\cite{Matterport3D}, Stanford2D3D~\cite{armeni2017joint}, 3D60~\cite{zioulis2018omnidepth}, and PanoSUNCG~\cite{wang2018self}. Matterport3D and Stanford2D3D are real-world datasets collected by Matterport’s Pro 3D Camera. While Matterport3D provides the raw depth, Stanford2D3D constructs the depth maps from reconstructed 3D models. Thus, the bottom and top depth is missed in Matterport3D, and some depth in Stanford2D3D is inaccurate, as shown in Fig.~\ref{fig:comparison1}. 3D60 is a $360^{\circ}$ depth dataset provided by Omnidepth \cite{zioulis2018omnidepth}, and it is rendered from 3D models of two realistic datasets, Matterport3D and Stanford2D3D, and two synthetic datasets, SceneNet~\cite{handa2016scenenet} and SunCG~\cite{song2017semantic}. In contrast, PanoSUNCG is a purely virtual dataset rendered from SunCG~\cite{song2017semantic}. The statistics of the $4$ datasets are listed in Tab.~\ref{tab:datasets}, and real datasets are smaller than virtual ones.

\subsubsection{Implementation Details}
We implement the proposed approach using Pytorch~\cite{paszke2017automatic}. 
The ResNet18~\cite{he2016deep} pretrained on ImageNet~\cite{deng2009imagenet} is used as backbone for most experiments, except for some using other backbones in Sec.~\ref{sec:ca}.
We use Adam~\cite{kingma2014adam} with default parameters as the optimizer and a constant learning rate of $0.0001$. 
Besides the common data augmentation techniques, random color adjustment, and left-right-flipping, we also use the random yaw rotation, as the ERP property is invariant under such transformation. 
We use the popular BerHu loss \cite{laina2016deeper} as the regression objective in training. 
During training, we randomly select 40 and 800 samples from the training set of Stanford2D3D and 3D60 for validation, and we use the last five scenes (1091 samples) from 80 training scenes of PanoSUNCG for validation. 
We train the real datasets for 100 epochs, and the virtual datasets for 30 epochs, as the virtual datasets are quite large, while BiFuse~\cite{wang2020bifuse} trains all datasets for 100 epochs. 
Following BiFuse, we set the input size for real and virtual datasets to $512\times1024$ and $256\times 512$. We train most models on an NVIDIA 2080Ti GPU, the batch size of virtual datasets is 8, but the batch size of real datasets is just 6 due to the limited GPU memory. {For some models in Sec.~\ref{sec:ca}, we have to use two GPUs, each with a batch size of 3. These models include UniFuse with CuP~\cite{cheng2018cube}, SP~\cite{wang2020bifuse} and CirP~\cite{wang2018omnidirectional}, and the equirectangular baseline with DaC~\cite{coors2018spherenet}.}

\subsubsection{Evaluation Metrics}
We use some standard metrics for evaluation, including four error metrics, mean absolute error (MAE), absolute relative error (Abs Rel), root mean square error (RMSE) and the root mean square error in log space (RMSE$\log$), and three accuracy metrics, \ie, the percentages of pixels where the ratio ($\delta$) between the estimated depth and ground truth depth is smaller than $1.25$, $1.25^2$, and $1.25^3$. Note that, while most papers on depth estimation use $\log_{e}$ in RMSE$\log$, the latest BiFuse adopts $\log_{10}$. As BiFuse is the state-of-the-art method with which we mainly compare our UniFuse model, we also adopt $\log_{10}$.

\begin{table}[t]
\vspace{5pt}
  \centering
  \resizebox{0.48\textwidth}{!}{
  \begin{tabular}{ l |cccc}
  \toprule
  {Dataset} & Matterport3D & Stanford2D3D & 3D60 & PanoSUNCG \\
  \hline
  \#train  & 7829 & 1040 & 35979 & 21025\\
  \#validation & 947 & - & - & - \\
  \#test & 2014 & 373 & 1298 & 3944\\
  \bottomrule
  \end{tabular}
  }
\caption{\textbf{The Statistics of the Datasets.}}
\label{tab:datasets}
\end{table}

\subsection{Experimental Results}

\subsubsection{Performance Comparison}
\label{sec:pc}
The quantitative comparison among the start-of-the-art methods of spherical depth estimation, our equirectangular baseline, and UniFuse model on the four datasets are shown in Tab.~\ref{tab:four}. We directly take the results from related papers for comparison. Our UniFuse model has established new state-of-the-art performance on all of the four datasets, especially on the biggest realistic dataset, Matterport3D, by a significant margin. To be specific, UniFuse outperforms BiFuse~\cite{wang2020bifuse} by reducing the error Abs Rel from 0.2048 to 0.1063 and improving accuracy metric of $\delta<1.25$ by $4.45\%$. In terms of fusion effectiveness, our UniFuse framework reduces the error metrics by over $10\%$ in average from our equirectangular baseline, while BiFuse only reduces by about $4\%$ from its equirectangular baseline. Although UniFuse's improvement on the loosest accuracy metric of $\delta<1.25^3$ is slightly smaller than BiFuse's improvement, UniFuse performs much better than BiFuse on enhancing the other two tighter accuracy metrics, especially over $3$ times on the accuracy of $\delta<1.25$.

\begin{table*}[t]
\vspace{5pt}
  \centering
  \resizebox{0.95\textwidth}{!}{
  \begin{tabular}{  c | l |cccc |ccc}
  \toprule
  \multirow{2}{*}{Dataset} & \multirow{2}{*}{Method} & \multicolumn{4}{c|}{Error metric $\downarrow	$} & \multicolumn{3}{c}{Accuracy metric $\uparrow	$}\\
  \cline{3-6}
  \cline{7-9}
   & & MAE & Abs Rel & RMSE  & RMSE$\log$ & $\delta < 1.25 $ & $\delta < 1.25^{2}$ & $\delta < 1.25^{3}$  \\

  \hline
  \hline
  \multirow{6}{*}{Matterport3D} &
  BiFuse~\cite{wang2020bifuse} --Equi. & 0.3701 & 0.2074  & 0.6536 & 0.1176 & 83.02 & 92.45 & 95.77 \\
  & BiFuse~\cite{wang2020bifuse} --Fusion & 0.3470  & 0.2048  &  0.6259 & 0.1134 & 84.52 & 93.19 & 96.32 \\
  &\cellcolor[gray]{.9}BiFuse~\cite{wang2020bifuse} --Improve &\cellcolor[gray]{.9}-6.24\% &\cellcolor[gray]{.9}-1.25\%  &\cellcolor[gray]{.9}-4.24\% &\cellcolor[gray]{.9}-3.57\% &\cellcolor[gray]{.9}+1.50 &\cellcolor[gray]{.9}+0.74 &\cellcolor[gray]{.9}\bf+0.55 \\
  \cline{2-9}
  & Ours --Equi. & 0.3267 & 0.1304  & 0.5460 & 0.0817 & 83.70 & 94.84 & 97.81 \\
  & Ours --Fusion & \bf0.2814  &\bf 0.1063  & \bf 0.4941 & \bf0.0701 & \bf88.97 & \bf96.23 & \bf98.31 \\
  &\cellcolor[gray]{.9}Ours --Improve &\cellcolor[gray]{.9}\bf-13.87\% &\cellcolor[gray]{.9}\bf-18.48\%  &\cellcolor[gray]{.9}\bf-9.51\% &\cellcolor[gray]{.9}\bf-14.20\% &\cellcolor[gray]{.9}\bf+5.27 &\cellcolor[gray]{.9}\bf+1.39 &\cellcolor[gray]{.9}+0.50\\
  
  \hline
  \hline
  \multirow{7}{*}{Stanford2D3D} &
  Jin \etal~\cite{jin2020geometric} & - & 0.1180  & 0.4210 & - & 85.10 & \bf 97.20 & 98.60 \\
  \cline{2-9}
  &BiFuse~\cite{wang2020bifuse} --Equi. & 0.2711 & 0.1428 & 0.4637 & 0.0911 & 82.61 & 94.58 & 98.00 \\
  &BiFuse~\cite{wang2020bifuse} --Fusion & 0.2343  & 0.1209  &  0.4142 & 0.0787 & 86.60 & 95.80 & 98.60 \\
  &\cellcolor[gray]{.9}BiFuse~\cite{wang2020bifuse} --Improve &\cellcolor[gray]{.9}-13.57\% &\cellcolor[gray]{.9}-15.34\%  &\cellcolor[gray]{.9} -10.68\% &\cellcolor[gray]{.9} -13.61\% &\cellcolor[gray]{.9}+3.99 &\cellcolor[gray]{.9}\bf +1.22 &\cellcolor[gray]{.9}\bf +0.60 \\
  \cline{2-9}
  &Ours --Equi. & 0.2696 & 0.1417 & 0.4224 & 0.0871 & 82.96 & 95.59 & 98.35 \\
  &Ours --Fusion & \bf 0.2082  & \bf 0.1114  & \bf 0.3691 & \bf 0.0721 & \bf 87.11 & 96.64 & \bf 98.82 \\
  &\cellcolor[gray]{.9}Ours --Improve &\cellcolor[gray]{.9}\bf -22.77\% &\cellcolor[gray]{.9}\bf -21.38\%  &\cellcolor[gray]{.9}\bf -12.62\% &\cellcolor[gray]{.9}\bf -17.22\% &\cellcolor[gray]{.9}\bf +4.15 &\cellcolor[gray]{.9} +1.05 &\cellcolor[gray]{.9} +0.47 \\

  \hline
  \hline

\multirow{8}{*}{3D60} &
  OmniDepth~\cite{zioulis2018omnidepth} & - & 0.0702  & 0.2911 & 0.1017\textdagger & 95.74 & 99.33 & 99.79 \\
  &Cheng \etal~\cite{cheng2020ode} & - & 0.0467  & \bf 0.1728 & 0.0793\textdagger & 98.14 & \bf 99.67 & \bf 99.89 \\
  \cline{2-9}
  &BiFuse~\cite{wang2020bifuse} --Equi. & 0.1172 & 0.0606 & 0.2667 & 0.0437 & 96.67 & 99.20 & 99.66 \\
 &BiFuse~\cite{wang2020bifuse} --Fusion & 0.1143  & 0.0615  &  0.2440 & 0.0428 & 96.99 & 99.27 & 99.69 \\
  &\cellcolor[gray]{.9}BiFuse~\cite{wang2020bifuse} --Improve &\cellcolor[gray]{.9}-2.47\% &\cellcolor[gray]{.9}+1.49\%  &\cellcolor[gray]{.9}\bf  -8.51\% &\cellcolor[gray]{.9}-2.06\% &\cellcolor[gray]{.9}+0.32 &\cellcolor[gray]{.9}\bf  +0.07 &\cellcolor[gray]{.9}\bf  +0.03 \\
  \cline{2-9}
  &Ours --Equi. &  0.1099 & 0.0517 & 0.2134  & 0.0342 & 97.64 & 99.59 & 99.86 \\

  &Ours --Fusion & \bf 0.0996  & \bf 0.0466  &  0.1968 & \bf 0.0315 &\bf 98.35 & 99.65 & 99.87 \\

  &\cellcolor[gray]{.9}Ours --Change &\cellcolor[gray]{.9}\bf -9.37\% &\cellcolor[gray]{.9}\bf  -9.86\%  &\cellcolor[gray]{.9}-7.78\% &\cellcolor[gray]{.9}\bf -8.16\% &\cellcolor[gray]{.9}\bf  +0.71 &\cellcolor[gray]{.9}+0.06 &\cellcolor[gray]{.9}+0.01 \\
  
  \hline
  \hline
  
  \multirow{6}{*}{PanoSUNCG} &
  BiFuse~\cite{wang2020bifuse} --Equi. & 0.0836 & 0.0687 & 0.2902 & 0.0496 & 95.29 & 97.87 & 98.86 \\
  &BiFuse~\cite{wang2020bifuse} --Fusion & 0.0789  & 0.0592  & \bf 0.2596 & 0.0443 & 95.90 & 98.38 & 99.07 \\
  &\cellcolor[gray]{.9}BiFuse~\cite{wang2020bifuse} --Improve &\cellcolor[gray]{.9}-5.62\% &\cellcolor[gray]{.9}\bf -13.83\%  &\cellcolor[gray]{.9}\bf -10.54\% &\cellcolor[gray]{.9}\bf -10.69\% &\cellcolor[gray]{.9}\bf +0.61 &\cellcolor[gray]{.9}\bf +0.51 &\cellcolor[gray]{.9}\bf +0.21 \\
  \cline{2-9}
  & Ours --Equi. & 0.0839 & 0.0531  & 0.2982 &  0.0444 & 96.09 & 98.25 & 99.00 \\
  & Ours --Fusion &\bf 0.0765  & \bf0.0485  & 0.2802 &\bf 0.0416 &\bf 96.55 &\bf 98.46 &\bf 99.12 \\
  &\cellcolor[gray]{.9}Ours --Improve &\cellcolor[gray]{.9}\bf -8.82\% &\cellcolor[gray]{.9}-8.66\%  &\cellcolor[gray]{.9}-6.04\% &\cellcolor[gray]{.9}-6.31\% &\cellcolor[gray]{.9}+0.46 &\cellcolor[gray]{.9}+0.21 &\cellcolor[gray]{.9}+0.12 \\
  
  \bottomrule
  \end{tabular}
  }
\caption{\textbf{Quantitative Comparison on Four Datasets.} \textdagger The RMSE$\log_{e}$ of our UniFuse on 3D60 is 0.0725.}
\label{tab:four}
\end{table*}

For the Stanford2D3D dataset, UniFuse outperforms BiFuse and another recent method by Jin~\etal~\cite{jin2020geometric}. UniFuse performs the best on most metrics except being sightly inferior to the model by Jin~\etal~\cite{jin2020geometric} on $\delta<1.25^2$. Note that, Jin~\etal~only experimented on a small portion of the Stanford2D3D dataset that satisfies the Manhattan structure (404 samples of the training set and 113 of the test set), as they proposed joint learning of $360^{\circ}$ depth and Manhattan layout. In contrast, UniFuse is not limited to such a specific structure, and if the model by Jin~\etal \ were evaluated on the entire test set of Stanford2D3D, its performance might degrade largely. In terms of fusion effectiveness, both UniFuse and BiFuse~\cite{wang2020bifuse} reduce the errors on Stanford2D3D more largely than Matterport3D, perhaps because the former is much less complex. Similarly, UniFuse outperforms BiFuse to a bigger extent on the other four error metrics. For the accuracy metrics, our UniFuse still has a better improvement on the tightest one than BiFuse. 

For 3D60, our UniFuse still has a much better improvement on MAE, Abs Rel, RMSElog, and $\delta<1.25$ and slightly less improvement on other three metrics than BiFuse~\cite{wang2020bifuse}. Overall, the performance on 3D60 is much higher than the two realistic datasets, and thus the effectiveness of the fusion is inferior. Our UniFuse model significantly outperforms BiFuse and performs approximately to the model by Cheng~\etal~\cite{cheng2020ode}. However, their model takes the depth map of the front face and extends it to the entire $360^{\circ}$ space. This requires an extra depth camera and careful calibration between the depth camera and panorama camera. 
The virtual PanoSUNCG is also very easy, and our final model still outperforms BiFuse at most metrics expect the RMSE.

\begin{figure*}[!ht]
\vspace{5pt}
\centering
\resizebox{0.96\textwidth}{!}{
\newcommand{\turnheightnew}{0.176\columnwidth}

\centering

\renewcommand{\arraystretch}{0.5}
\begin{tabular}{@{\hskip 1mm} c@{\hskip 1mm}c@{\hskip 1mm}c@{\hskip 1mm}c@{\hskip 1mm}c@{}}

{\rotatebox{90}{\hspace{0.8mm}\scriptsize Matterport3D}} &
\includegraphics[height=\turnheightnew]{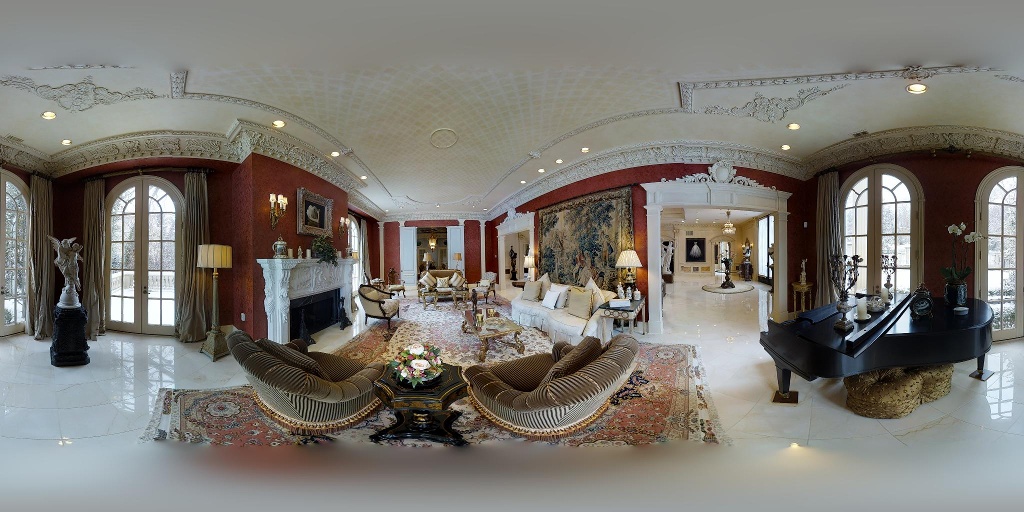} &
\includegraphics[height=\turnheightnew]{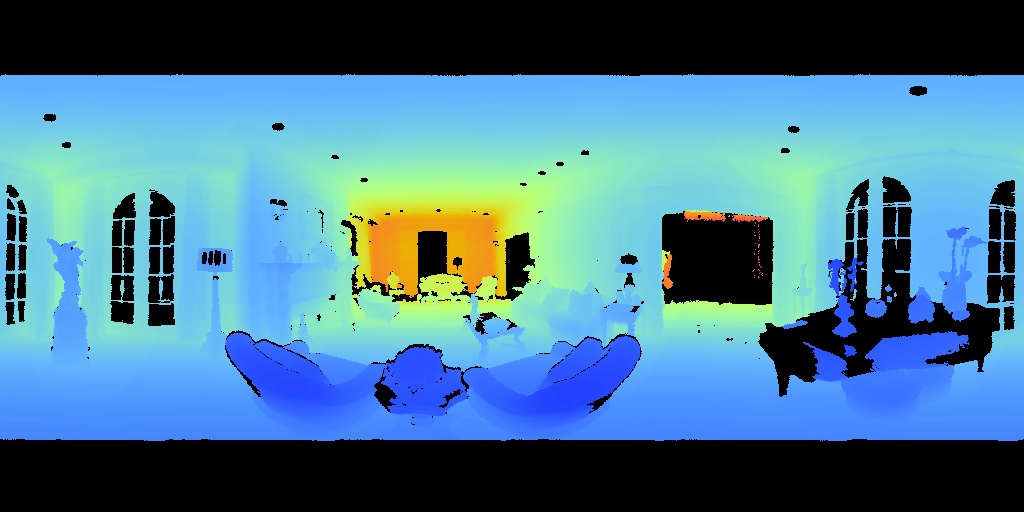} &
\includegraphics[height=\turnheightnew]{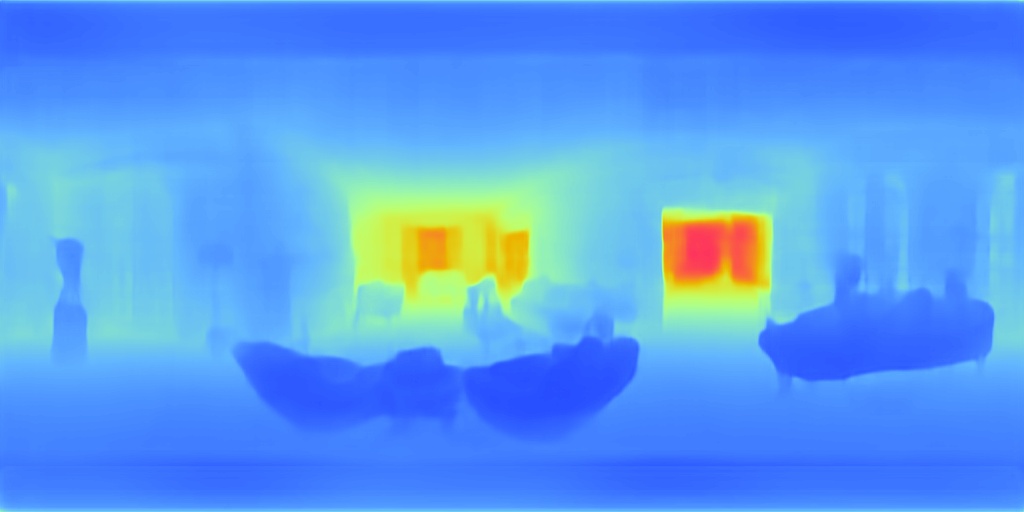} &
\includegraphics[height=\turnheightnew]{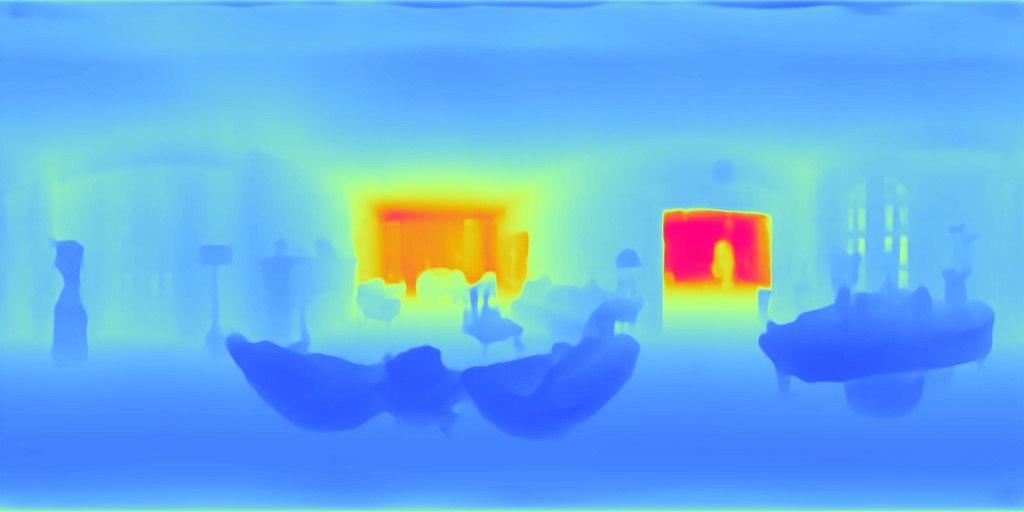}\\

{\rotatebox{90}{\hspace{0.8mm}\scriptsize Matterport3D}} &
\includegraphics[height=\turnheightnew]{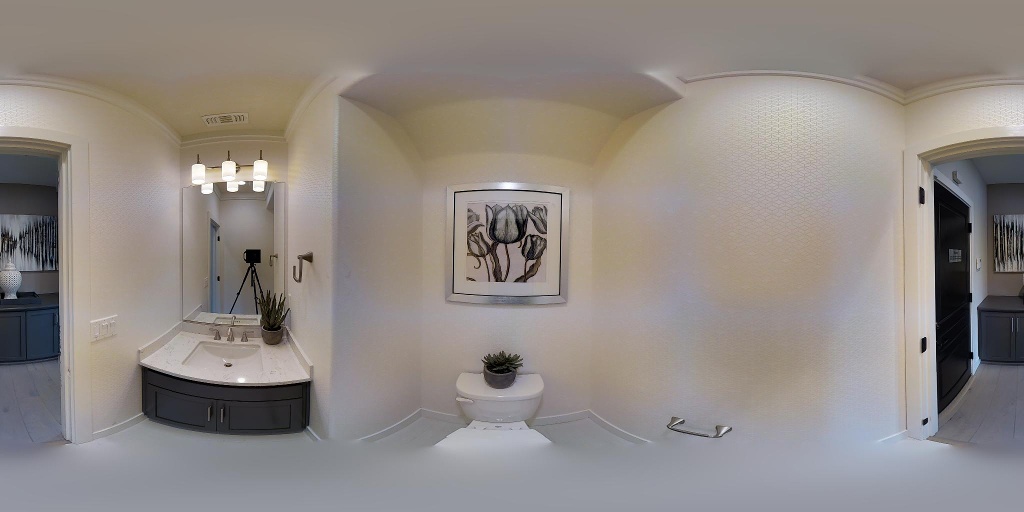} &
\includegraphics[height=\turnheightnew]{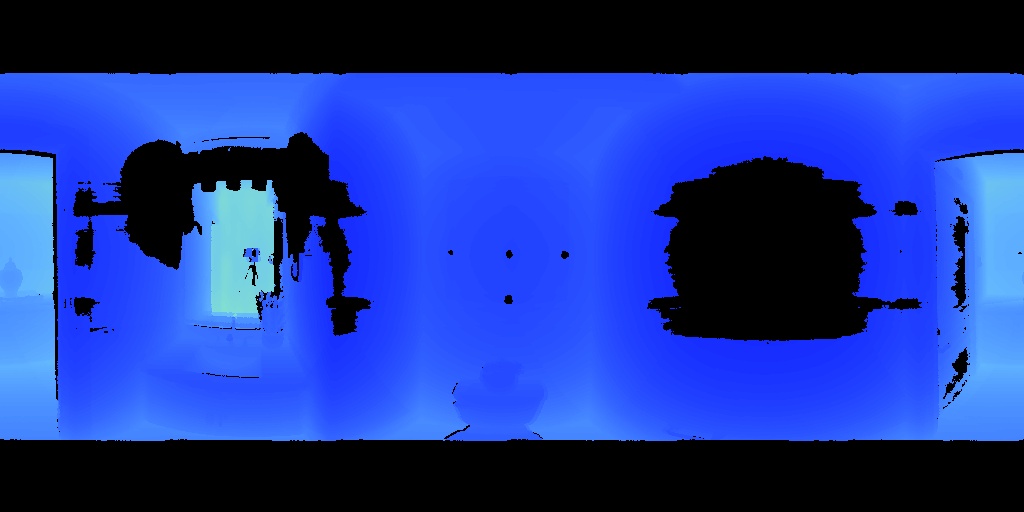} &
\includegraphics[height=\turnheightnew]{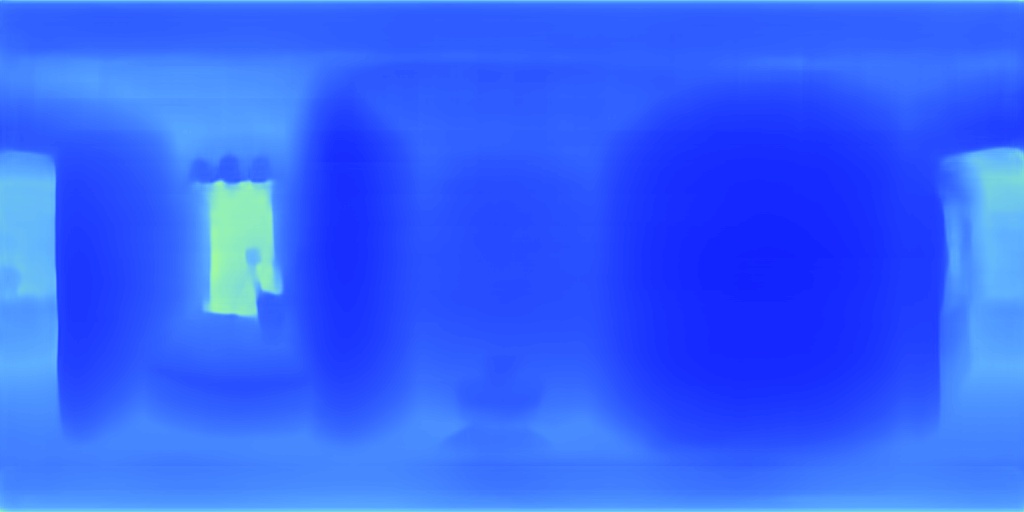} &
\includegraphics[height=\turnheightnew]{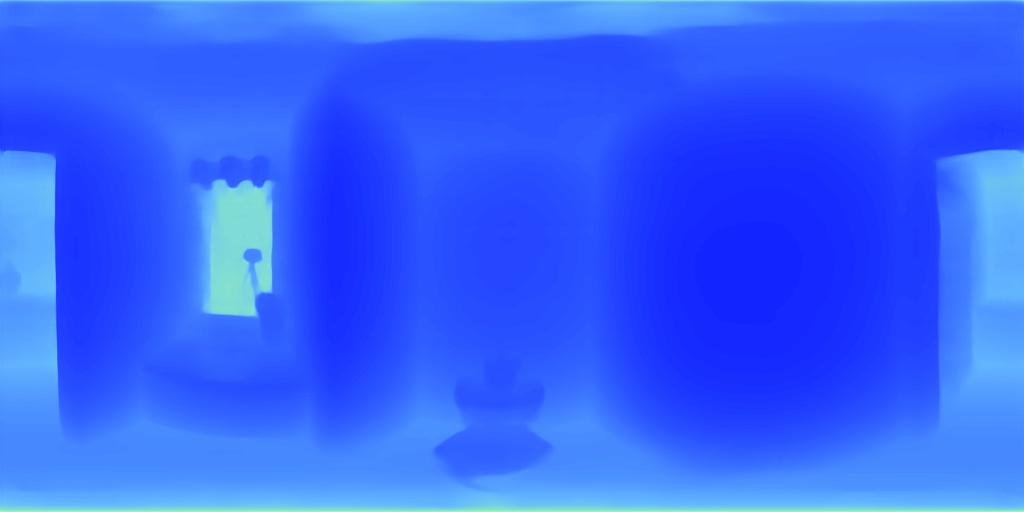}\\

{\rotatebox{90}{\hspace{0.1mm}\scriptsize Stanford2D3D}} &
\includegraphics[height=\turnheightnew]{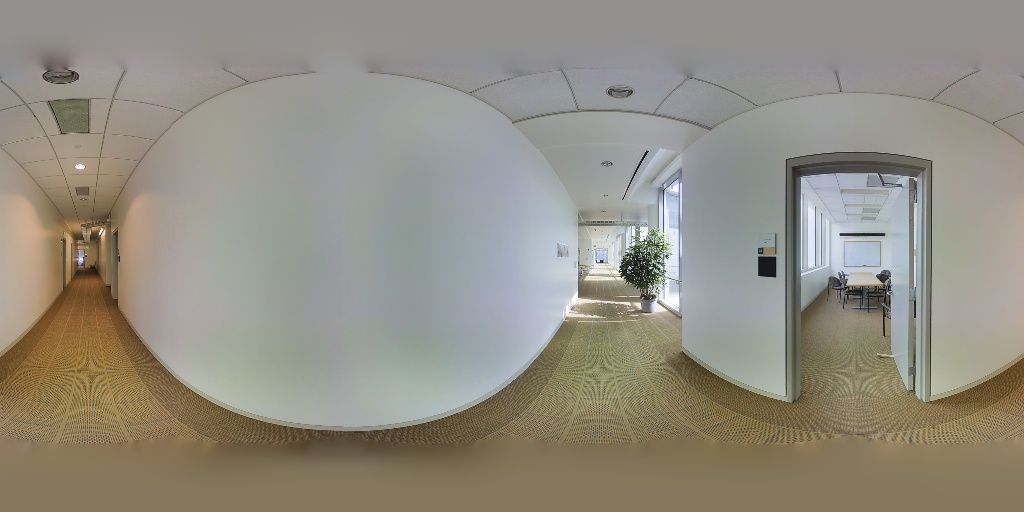} &
\includegraphics[height=\turnheightnew]{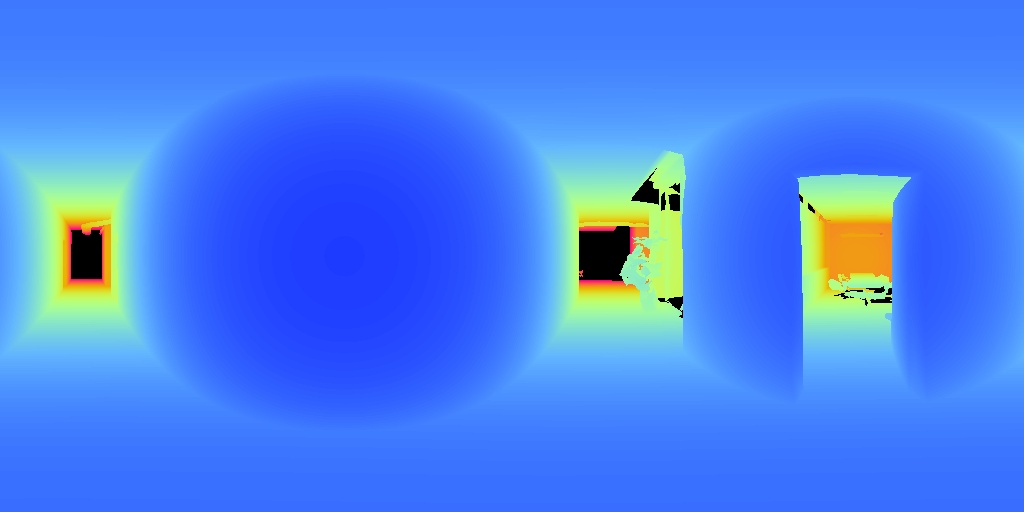} &
\includegraphics[height=\turnheightnew]{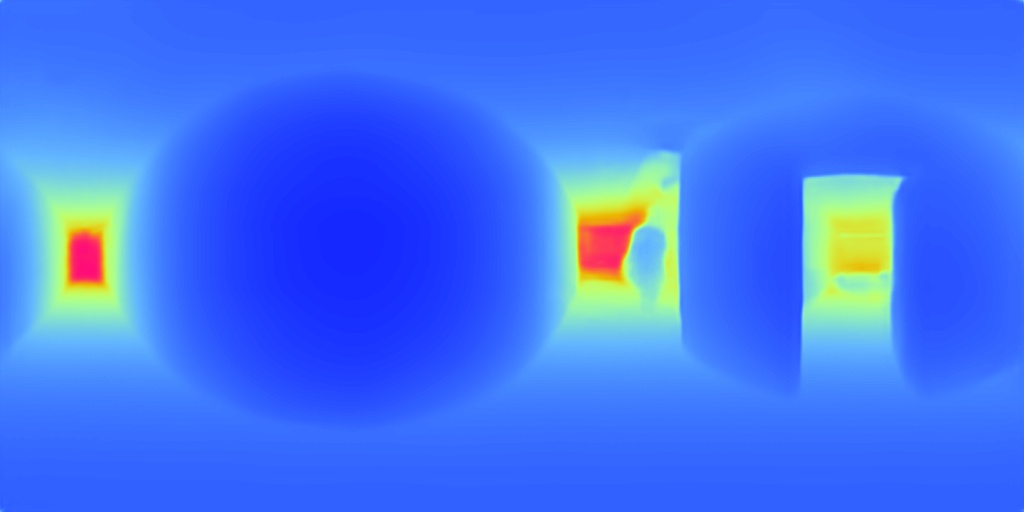} &
\includegraphics[height=\turnheightnew]{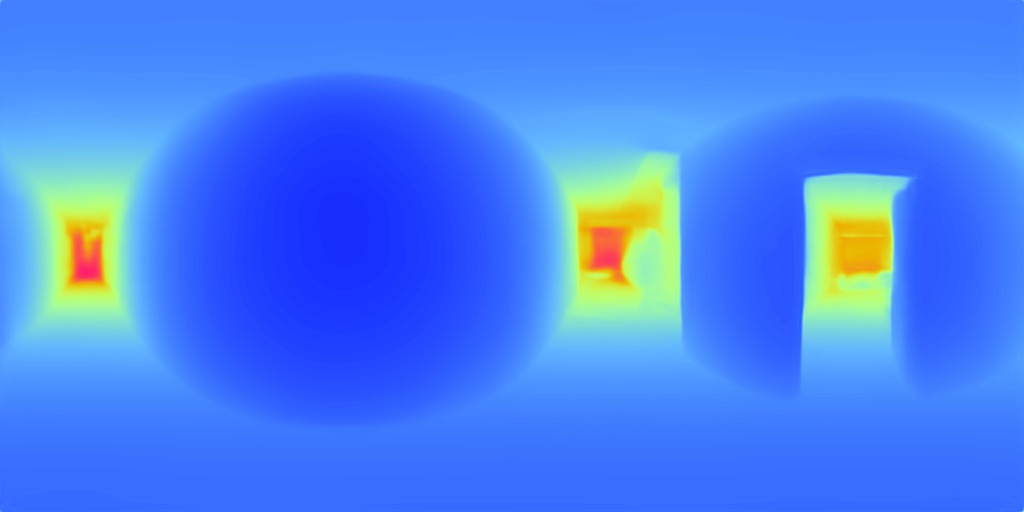}\\

{\rotatebox{90}{\hspace{0.1mm}\scriptsize Stanford2D3D}} &
\includegraphics[height=\turnheightnew]{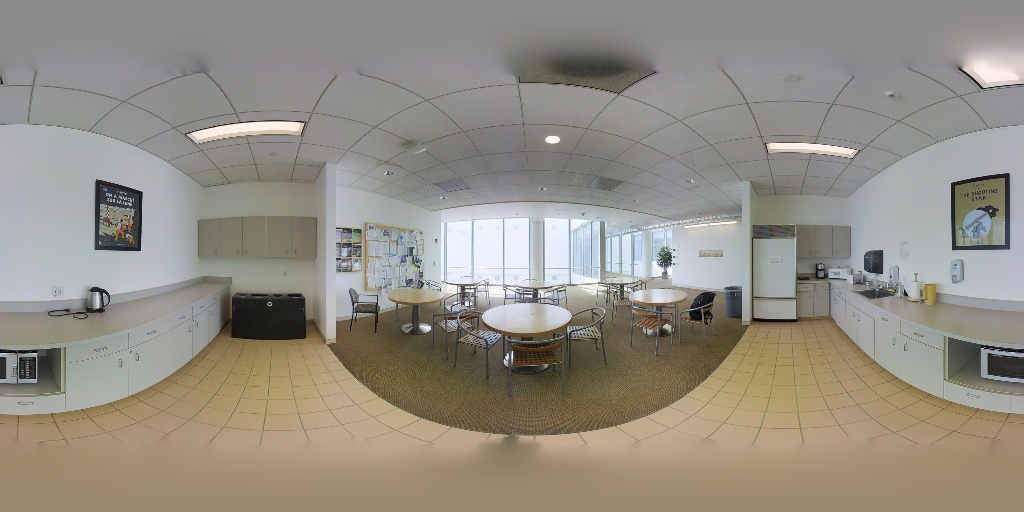} &
\includegraphics[height=\turnheightnew]{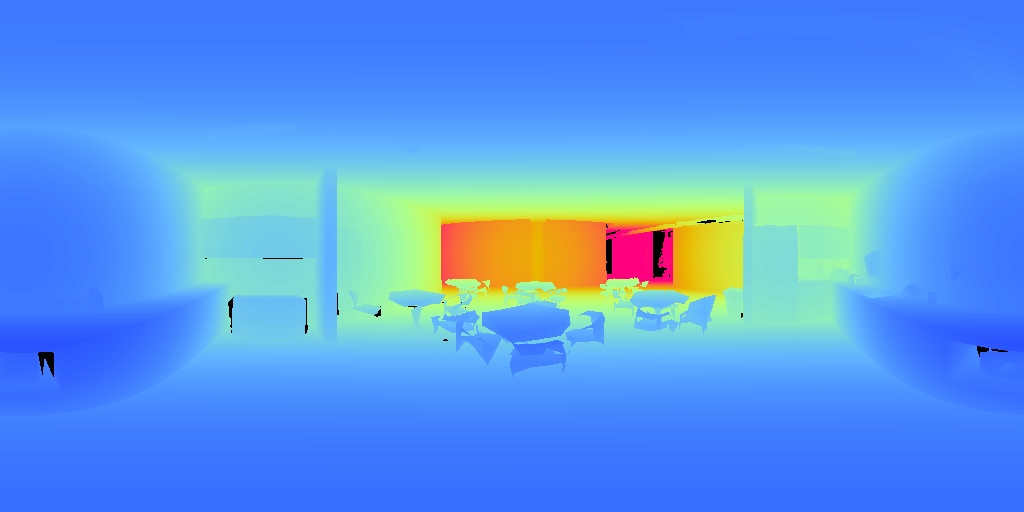} &
\includegraphics[height=\turnheightnew]{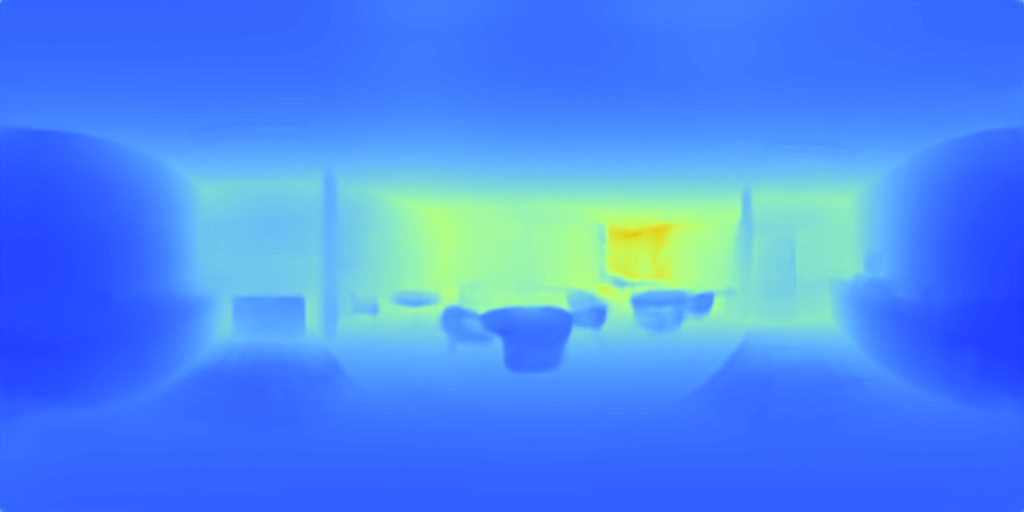} &
\includegraphics[height=\turnheightnew]{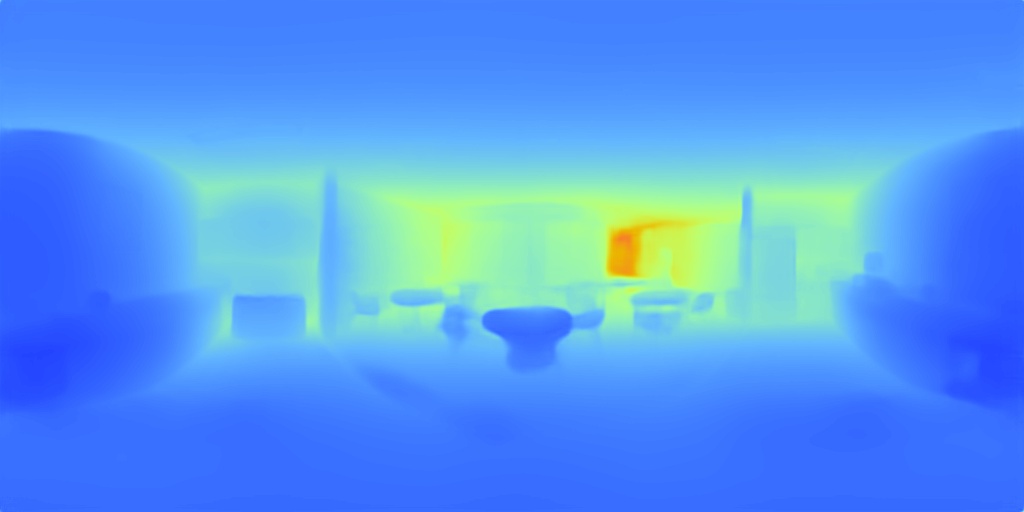}\\

{\rotatebox{90}{\hspace{5mm}\scriptsize 3D60}} &
\includegraphics[height=\turnheightnew]{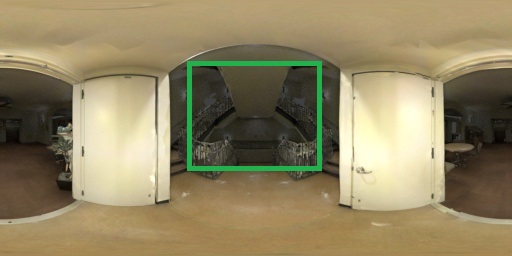} &
\includegraphics[height=\turnheightnew]{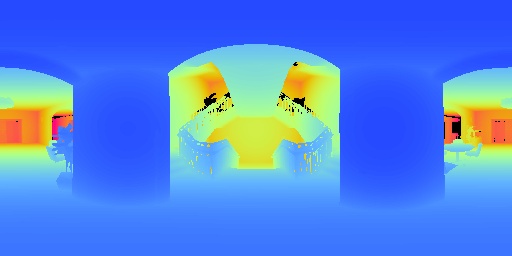} &
\includegraphics[height=\turnheightnew]{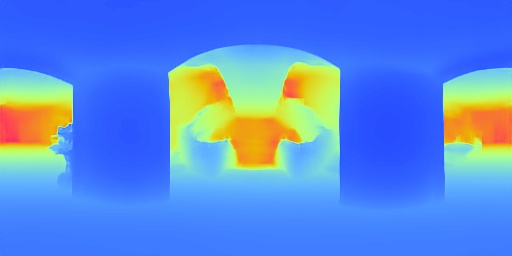} &
\includegraphics[height=\turnheightnew]{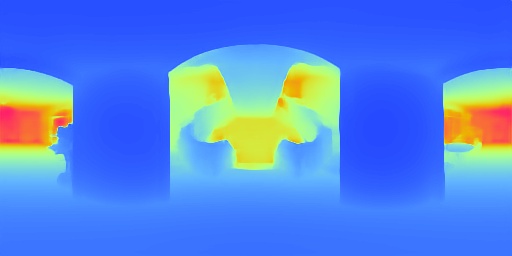}\\

{\rotatebox{90}{\hspace{5mm}\scriptsize 3D60}} &
\includegraphics[height=\turnheightnew]{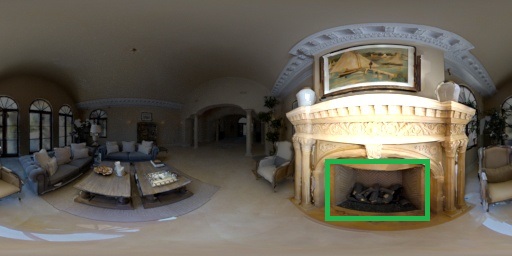} &
\includegraphics[height=\turnheightnew]{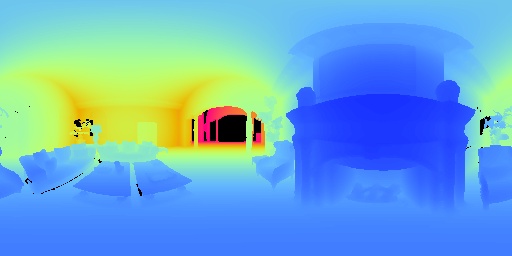} &
\includegraphics[height=\turnheightnew]{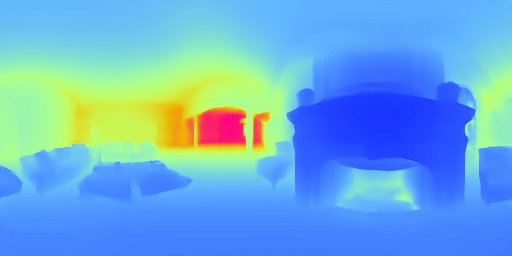} &
\includegraphics[height=\turnheightnew]{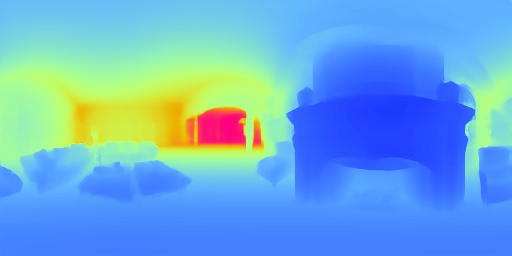}\\

{\rotatebox{90}{\hspace{1mm}\scriptsize PanoSUNCG}} &
\includegraphics[height=\turnheightnew]{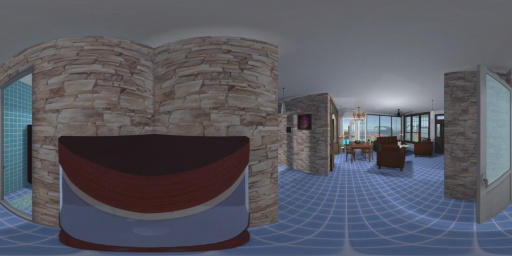} &
\includegraphics[height=\turnheightnew]{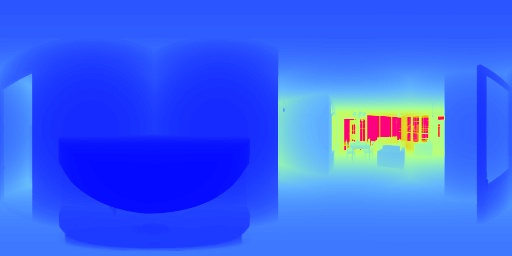} &
\includegraphics[height=\turnheightnew]{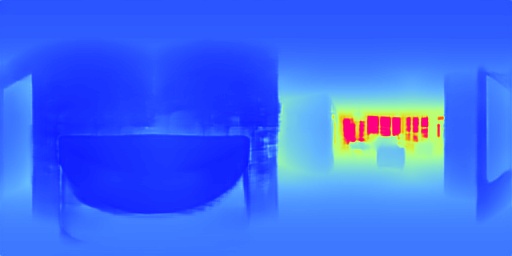} &
\includegraphics[height=\turnheightnew]{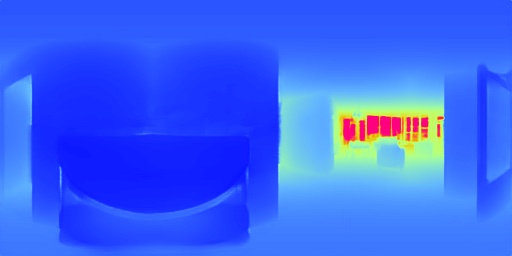}\\

{\rotatebox{90}{\hspace{1mm}\scriptsize PanoSUNCG}} &
\includegraphics[height=\turnheightnew]{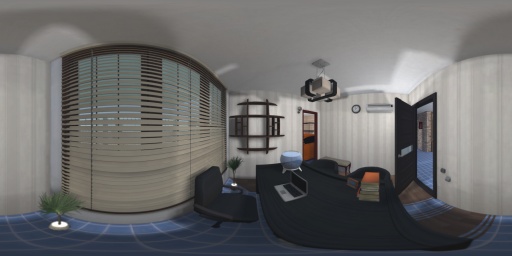} &
\includegraphics[height=\turnheightnew]{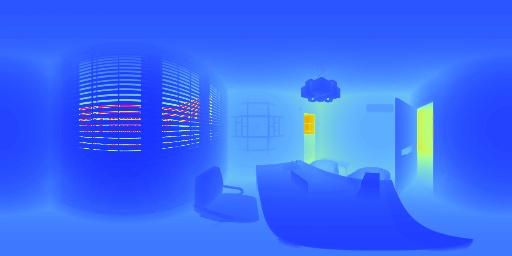} &
\includegraphics[height=\turnheightnew]{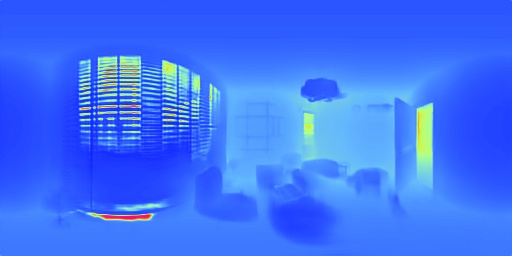} &
\includegraphics[height=\turnheightnew]{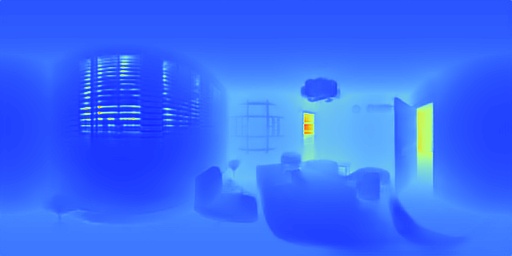}\\

&
\scriptsize Input &
\scriptsize Ground Truth&
\scriptsize Our Equi. &
\scriptsize Our UniFuse\\

\end{tabular}
 }
\caption{\textbf{Qualitative Comparison between Our Equirectangular Baseline and UniFuse Model.} Best viewed in color.}
\label{fig:comparison1}
\end{figure*}

We also provide qualitative results of our equirectangular baseline and UniFuse model in Fig.~\ref{fig:comparison1}. Two examples from the test set of each dataset are shown here and the dark region of the ground truth depth maps indicates unavailable depth. It can be observed that the UniFuse model produces accurate depth maps with fewer artifacts than the equirectangular baseline, which further verifies the effectiveness of our proposed approach. 
The two examples of the 3D60 dataset are rendered from the 3D models of Matterport3D, and it appears that the farther it gets, the darker the scene is. But in the realistic dataset, it is not the case. We believe that the problematic rendering makes 3D60 easy, as the network probably uses the brightness as a cue for predicting depth. However, sometimes such a cue may cause problems. For instance, the regions within a blue rectangle on the two examples contain dark areas, and our equirectangular baseline tends to predict the regions farther. 

\subsubsection{Ablation Study}

\begin{table}[t]
\vspace{6pt}
  \centering
  \resizebox{0.99\columnwidth}{!}{
  \begin{tabular}{ l |cccc}
  \toprule
  {Method} & MAE $\downarrow$ & Abs Rel $\downarrow$ & RMSE $\downarrow$ & $\delta < 1.25 \uparrow $  \\
  \hline
  {Equi. w/o pt} & {0.3548} & {0.1413} & {0.5946} & {81.53}    \\
  Equi. & 0.3267 & 0.1304 & 0.5460 & 83.70    \\
  \hline
  Concat. & 0.3162 & 0.1237 & 0.5340 & 85.00 \\
  Bi-Proj. &  0.3096 & 0.1188 & 0.5283 & 85.94   \\
  CEE w/o SE &  0.3046 & 0.1161 & 0.5217 & 86.53 \\
  \hline 
  {UniFuse w/o pt}& {0.3164} & {0.1195} & {0.5440} & {85.53}    \\
  UniFuse &  0.2814 & 0.1063 & 0.4941 & 88.97  \\
  \bottomrule
  \end{tabular}
  }
\caption{\textbf{Ablation Study.}}
\label{tab:ablation}
\end{table}

We compare the effectiveness of { ImageNet~\cite{deng2009imagenet} pretraining (pt) and} the three different fusion modules in our proposed fusion framework on the Matterport3D dataset in Tab.~\ref{tab:ablation}. 
{The pretraining is useful for both the baseline and UniFuse. For the baseline, disabling pretraining makes the Abs Rel error increase by $8.36\%$ and the $\delta<1.25$ accuracy drop over $2\%$. Disabling pretraining in both equirectangular and cubemap branches of  UniFuse results in a bigger performance degradation. Therefore, pretraining is also useful for ERP. There is an explanation for this effect in~\cite{jin2020geometric}, which adopts a pretrained U-Net for ERP. The reason is that the high-level parameters from perspective images can be more easily fine-tuned to the equirectangular ones.} 

\begin{table*}[t]
\vspace{5pt}
  \centering
  \resizebox{0.65\textwidth}{!}{
  \begin{tabular}{ l |ccc|cc}
  \toprule
  {Method} & \#Para.  & Mem. & Time & Abs Rel $\downarrow$  & $\delta < 1.25 \uparrow $  \\
  \hline
  \hline
  BiFuse's Euqi.~\cite{wang2020bifuse} (R50) & 63.56M & {2247M} & {113ms} & 0.2075 & 83.02 \\
  BiFuse~\cite{wang2020bifuse} (R50) & 253.1M & {4003M} & {1125ms} & 0.2048 & 84.52 \\
  
  \hline
  \hline
  Our Euqi. (R18) &  14.33M  & {907M}  & {12.3ms} & 0.1304 & 83.70 \\
  Our Euqi. (R50) &  32.52M  & {1039M} & {27.2ms} & 0.1207 & 85.31 \\

  UniFuse (R18)  & {30.26M} & {1221M} & {24.1ms} & 0.1063 & 88.97 \\
  \hline
  Our Euqi. (MV2)  &  4.00M  & {791M} & {13.9ms} & 0.1274 & 84.27 \\
  UniFuse (MV2)  & {7.35M} & {889M} & {28.9ms} & 0.1116 & 87.56 \\
  
  \hline
  \hline
  {UniFuse  w/ CuP~\cite{cheng2018cube} (R18)}  & {30.26M} & {1271M} & {365ms} & {0.1081} & {88.44} \\
  
  {UniFuse  w/ SP~\cite{wang2020bifuse} (R18)}  & {30.26M} & {1271M} & {574ms} & {0.1089} & {88.36} \\
  
  {Our Euqi. w/ CirP~\cite{wang2018omnidirectional} (R18)} &  {14.33M}  & {999M}  & {25.6ms} & {0.1229} & {85.39} \\
  
  {UniFuse  w/ CirP~\cite{wang2018omnidirectional} (R18)}  & {30.26M} & {1275M} & {40.8ms} & {0.1060} & {88.92} \\
  \hline
  {Our Euqi. w/ DaC~\cite{coors2018spherenet} (R18)} &  {14.33M}  & {1455M}  & {116ms} & {0.1194} & {86.00} \\
  
  {Our Euqi. w/ SC~\cite{su2017learning} (R18)} &  {14.34M}  & {897M}  & {12.6ms} & {0.1300} & {83.98} \\
  
  {UniFuse  w/ SC~\cite{su2017learning} (R18)}  & {30.27M} & {1239M} & {24.5ms} & {0.1134} & {87.45} \\
  \bottomrule
  \end{tabular}
  }
\caption{\textbf{Model Complexity Comparison.} }
\label{tab:complexity}
\vspace{-5pt}
\end{table*}

The simple concatenation of equirectangular features and cubemap features produces a good performance gain upon the equirectangular baseline. Specifically, the Abs Rel error is reduced by $5.14\%$, and the  $\delta<1.25$ accuracy increases $1.3\%$. 
This indicates that our unidirectional fusion strategy is effective, although simple. 
Our fusion strategy is also compatible with the Bi-Projection module in BiFuse \cite{wang2020bifuse}. Bi-Projection roughly doubles the performance gain of the naive concatenation. 
By looking backwards to Tab.~\ref{tab:four}, it is easily to find that the performance gain of our simple unidirectional fusion scheme with Bi-Projection significantly surpasses that of the complex bidirectional scheme of BiFuse. This further verified that our simplified fusion scheme is even more effective. 
Furthermore, our proposed CEE fusion module produces a large performance gap over Bi-Projection. To be specific, our CEE module has an Abs Rel $10.52\%$ less than Bi-Projection, and the accuracy $\delta<1.25$ is $3.03\%$ higher. When turning off the SE block of our CEE module, the performance is still markedly better than Bi-Projection. Thus, our proposed CEE module is much more effective in fusing the equirectangular features and cubemap features than Bi-Projection under our unidirectional fusion scheme.

\subsubsection{Complexity Analysis}
\label{sec:ca}
We compare the complexity among the models of this work and the models in BiFuse~\cite{wang2020bifuse} as well as some of their performance on Matterport3D to understand how the performance is boosted by adding complexity. 
{We also examine the efficiency and effectiveness of some existing methods handling discontinuity and distortion for panoramas in our models, which are listed at the end of Sec~\ref{sec:rw}. }
The complexity metrics include the number of {neural} model parameters, the GPU memory, and computation time when the model infers an image with a size of $512\times1024$.
The experiment is performed on {an NVIDIA Titan Xp GPU}, and the computation time is averaged over 1000 images. {The results are listed in Tab~\ref{tab:complexity}.} R50 and R18 indicates that the backbone are ResNet-50 and ResNet-18 \cite{he2016deep}, respectively, while MV2 represents MobileNetV2 \cite{sandler2018mobilenetv2}.

We obtained the complexity of BiFuse from its open inference code. The models of BiFuse are much more complicated than ours. 
{Its baseline is much more complex than our baseline. 
It is an FCRN \cite{laina2016deeper} with the R50 backbone whose first layer is replaced with SC~\cite{su2017learning}. Ours is just a simpler U-Net using a pretrained R18 backbone. However, our baseline still performs slightly better. }
BiFuse's fusion scheme significantly increases the complexity. It almost quadruples the number of parameters. One fold of parameter comes from the cubemap branch. The Bi-Projection modules on both encoding and decoding stages and the final refinement module to fuse the depth maps from both the equirectangular and cubemap branches increase the neural parameters further.
BiFuse also increases the inference time {to about $10$} times. We believe that only part of it comes from the Bi-Projection modules and the final refinement module. Most of it probably can be attributed to SP~\cite{wang2020bifuse} on the cubemap branch{, which will be explained later when using SP and CuP~\cite{cheng2018cube} in UniFuse}.

In contrast, our UniFuse on ResNet-18 only doubles the complexity of parameters and time upon our baseline, but the performance boosts, and it is still real-time ($>30fps$). To prove that the performance of our UniFuse is not simply obtained by adding complexity, we also experiment with our baseline with ResNet-50, whose complexity is similar to our UniFuse on ResNet-18 but the performance is far behind. To explore the possibility to apply our models on mobile devices, we also experiment on the MobileNetV2 \cite{sandler2018mobilenetv2}. The results show that our UniFuse can still boost performance, even to a smaller extent than using ResNet-18. The inference time is still real-time, and the GPU memory and parameters decrease a great deal, which indicates our models have potential to be applied to mobile robots or AR/VR devices.   

{The remaining experiments are about using special padding and convolution methods in our models, and the results indicate it is unnecessary to adopt them in UniFuse. We find that both SP and CuP are implemented in BiFuse's code, so we introduce them in UniFuse. The results show the inference time increases over $10$ times for CuP and over $20$ times for SP. They have similar implementations. For each of $6$ cube faces, it has $4$ sides on the feature map to be sampled from adjacent cube faces, so there are $24$ loops for each padding. The padding should be performed on convolutions with kernel size bigger than $1\time1$. Therefore, the inference time greatly increases. SP uses interpolation in each loop, so its time increases further. However, they do not improve the performance of UniFuse. Although BiFuse's paper has an ablation study where they enhance the predictions of the cubemap branch, we hypothesize that they are less useful for fusion, especially for UniFuse, which uses CMP as a supplement. 
A similar observation can be seen in the experiments of CirP~\cite{wang2018omnidirectional}. CirP almost does not improve the performance of UniFuse but is effective for our baseline. }

{There is a small technical difference for DaC among different studies~\cite{tateno2018distortion, coors2018spherenet, fernandez2020corners}. We implement the version by Coors~\etal~\cite{coors2018spherenet} in our equirectangular baseline. DaC improves the performance of our baseline but is far inferior to UniFuse. However, as interpolation has to be performed densely in convolution, the resulted space and time complexity is much higher, which is in accordance with the experiments of CFL~\cite{ fernandez2020corners}. We replace the first layer of our models with BiFuse's implementation of SC~\cite{su2017learning}. The resulted complexity is almost the same, since only the first layer is changed. SC slightly improves the performance of our baseline but reduces the performance of UniFuse. Therefore, SC is not effective for our UniFuse. }

\begin{figure*}[!ht]
\vspace{6pt}
\centering
\resizebox{0.96\textwidth}{!}{
\newcommand{\turnheightnew}{0.176\columnwidth}

\centering

\renewcommand{\arraystretch}{0.5}
\begin{tabular}{@{\hskip 1mm} c@{\hskip 1mm}c@{\hskip 1mm}c@{\hskip 1mm}c@{\hskip 1mm}c@{}}

{\rotatebox{90}{\hspace{0.8mm}\scriptsize Matterport3D}} &
\includegraphics[height=\turnheightnew]{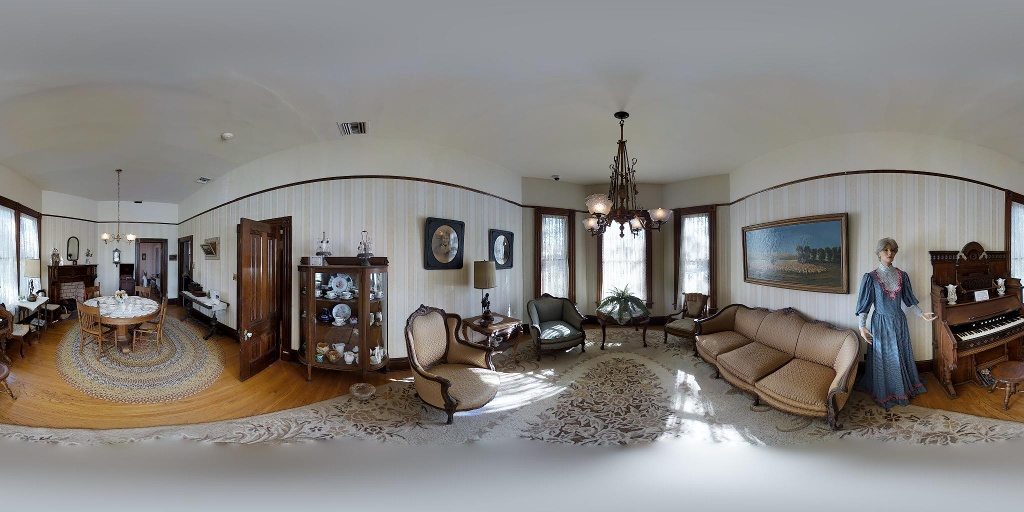} &
\includegraphics[height=\turnheightnew]{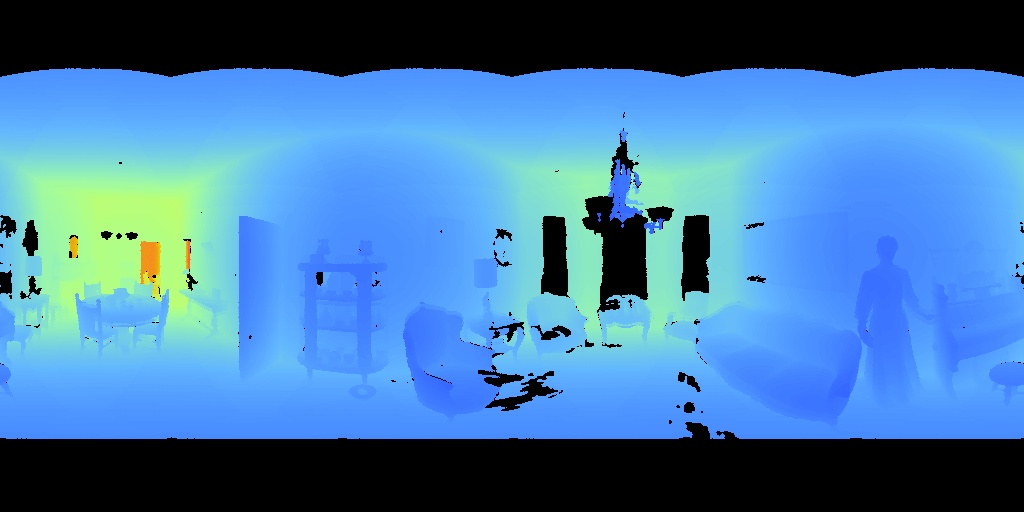} &
\includegraphics[height=\turnheightnew]{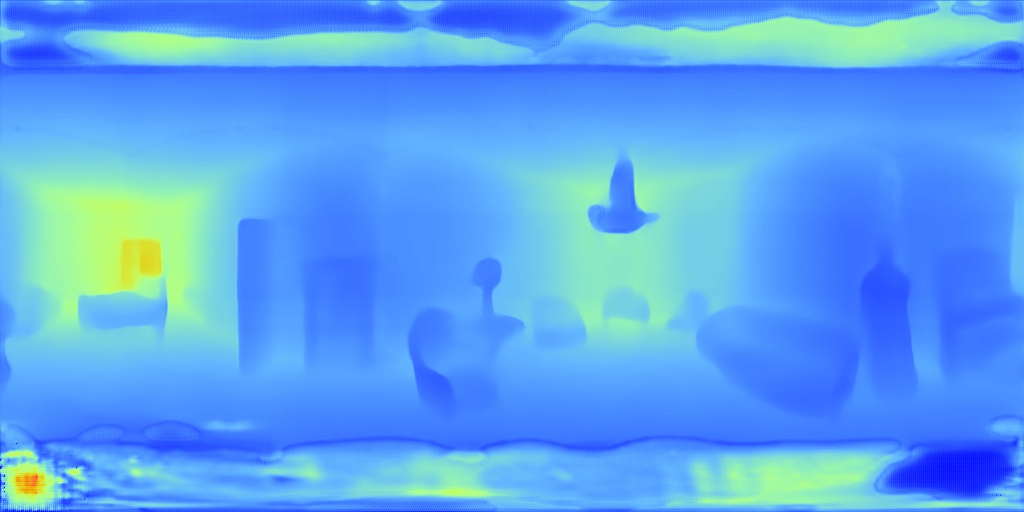} &
\includegraphics[height=\turnheightnew]{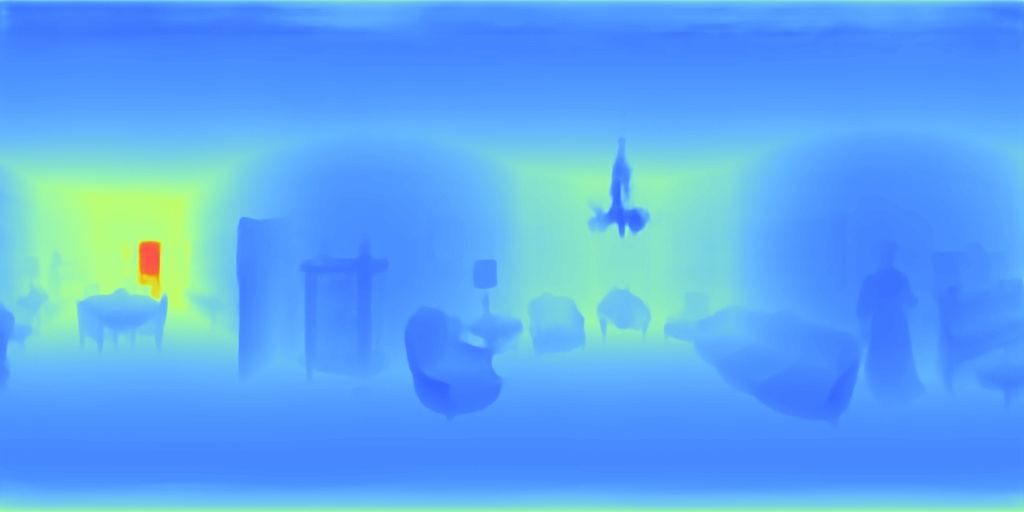}\\

{\rotatebox{90}{\hspace{0.8mm}\scriptsize Matterport3D}} &
\includegraphics[height=\turnheightnew]{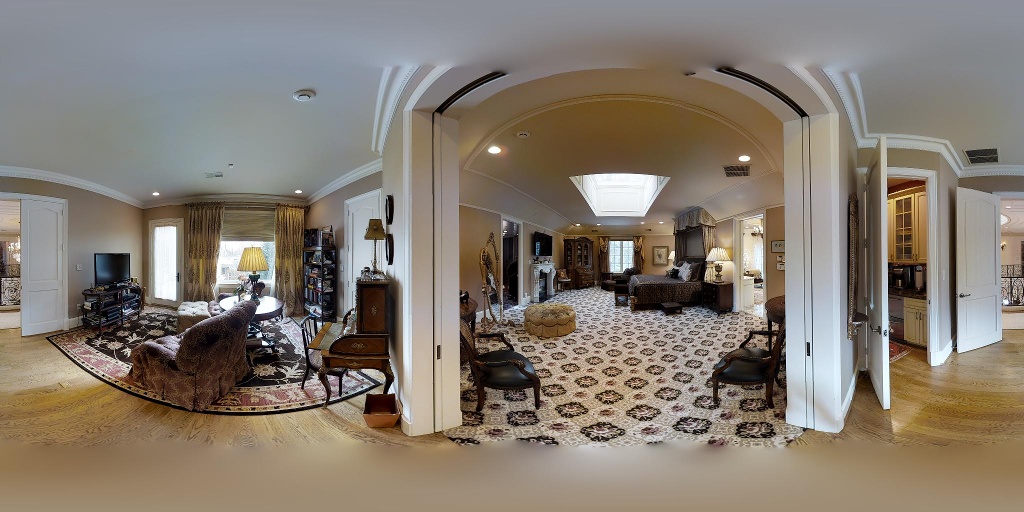} &
\includegraphics[height=\turnheightnew]{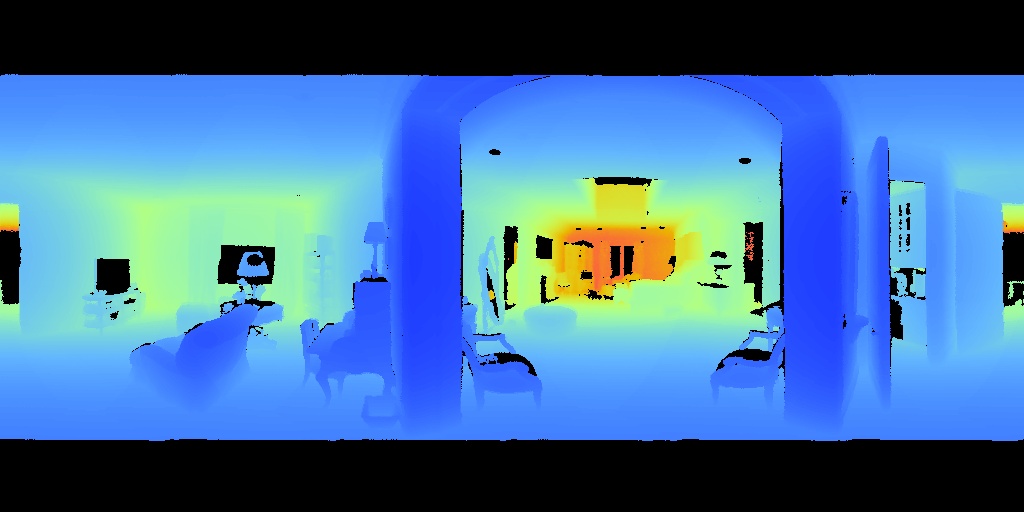} &
\includegraphics[height=\turnheightnew]{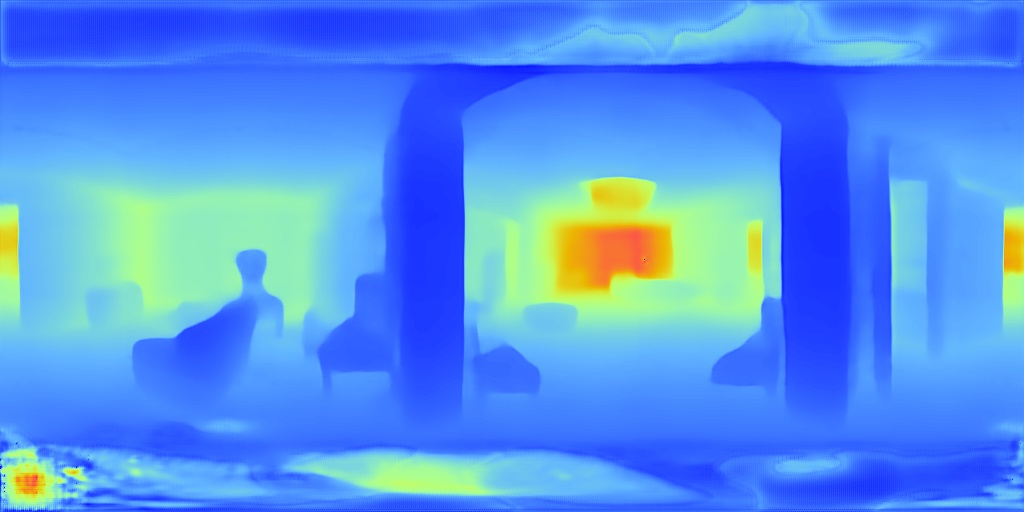} &
\includegraphics[height=\turnheightnew]{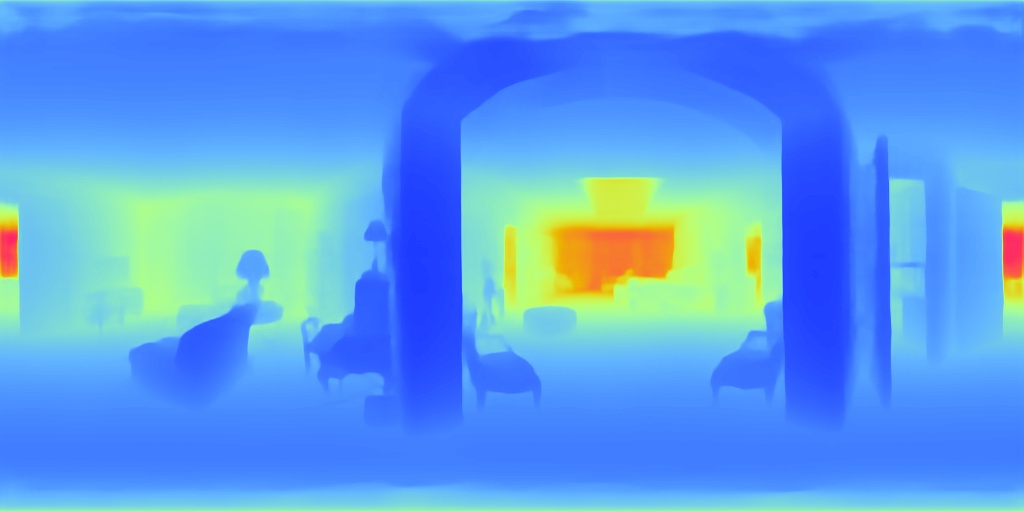}\\

{\rotatebox{90}{\hspace{0.8mm}\scriptsize Matterport3D}} &
\includegraphics[height=\turnheightnew]{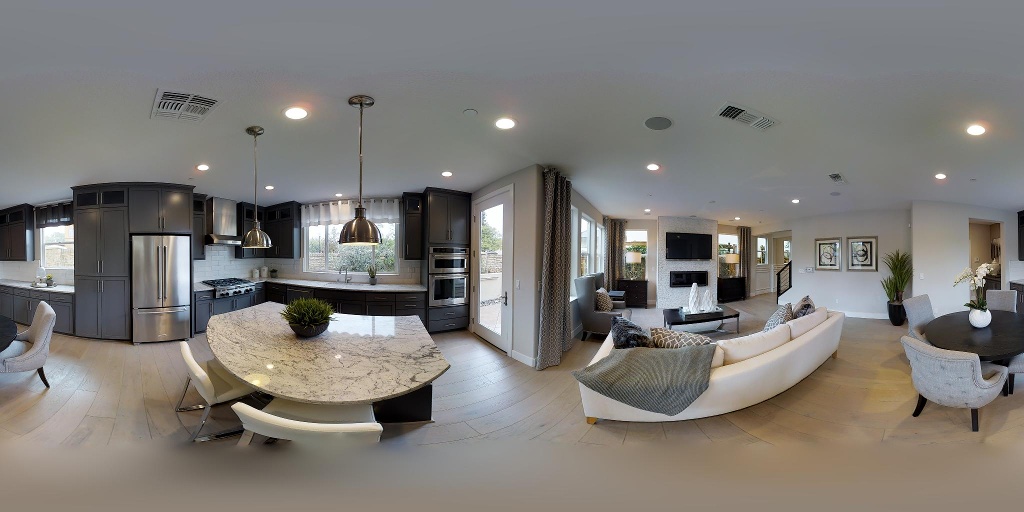} &
\includegraphics[height=\turnheightnew]{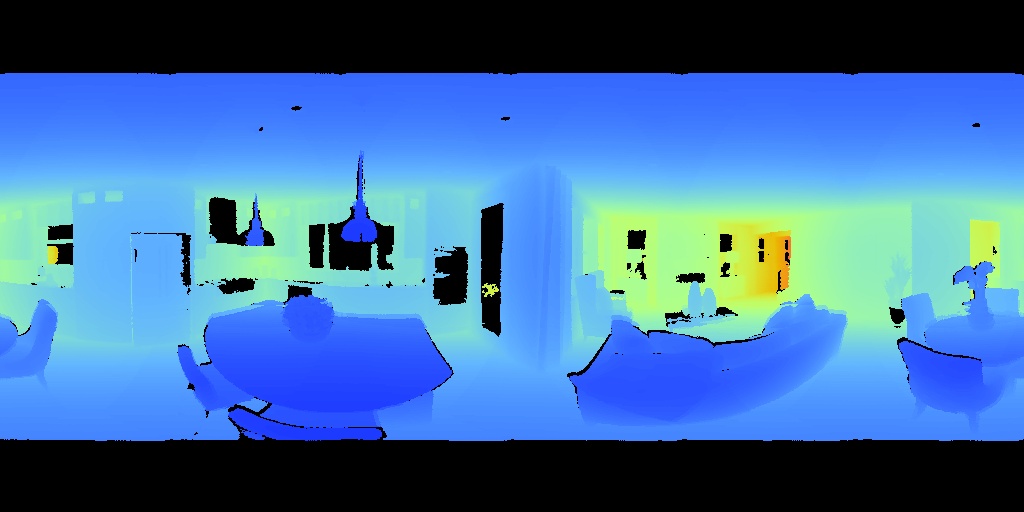} &
\includegraphics[height=\turnheightnew]{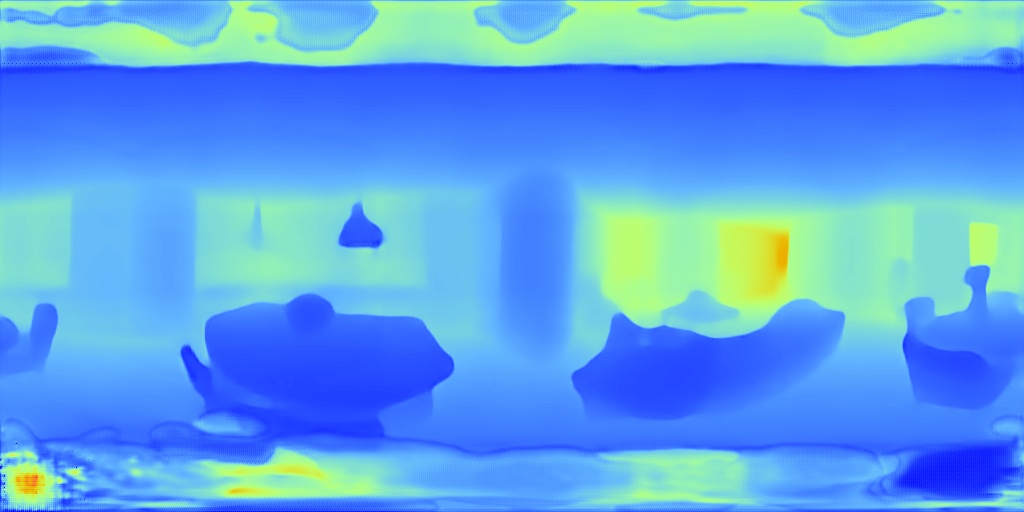} &
\includegraphics[height=\turnheightnew]{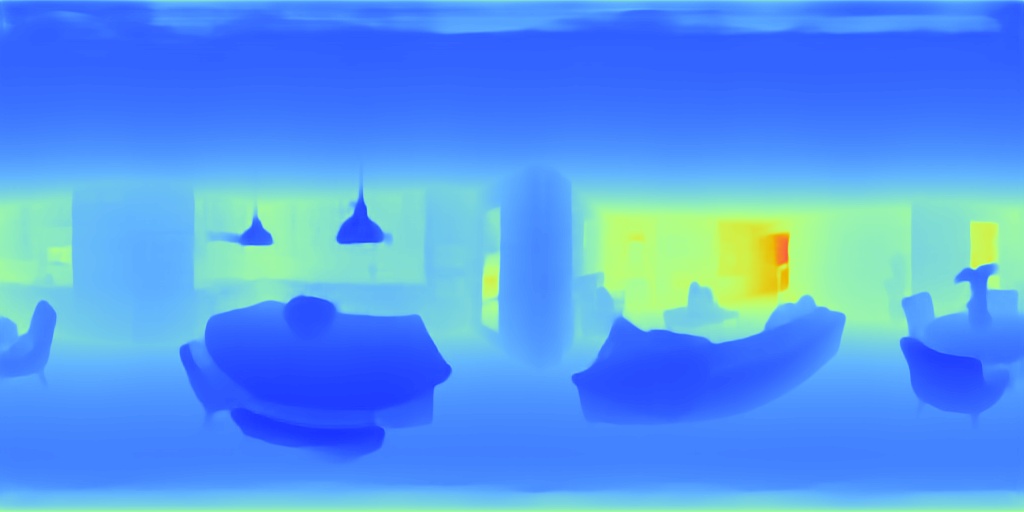}\\

{\rotatebox{90}{\hspace{0.1mm}\scriptsize Stanford2D3D}} &
\includegraphics[height=\turnheightnew]{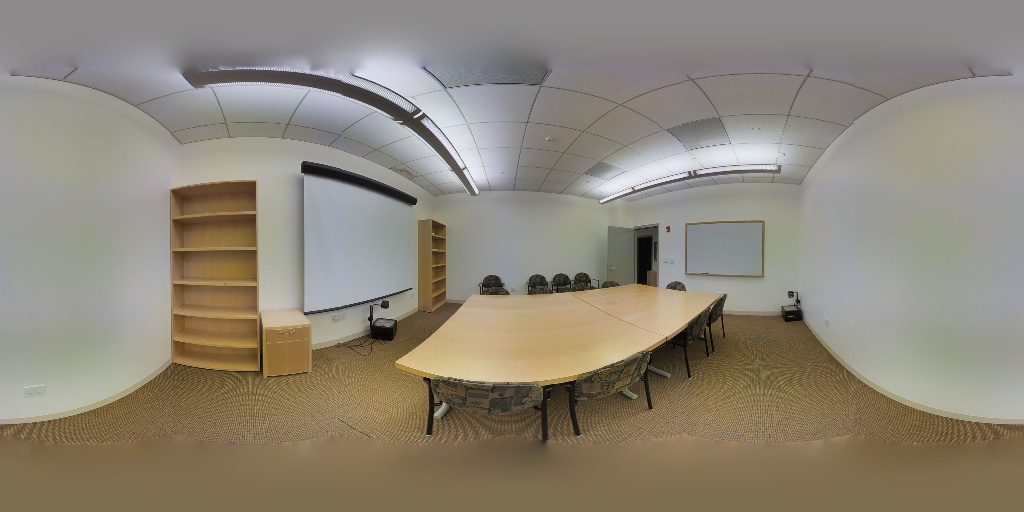} &
\includegraphics[height=\turnheightnew]{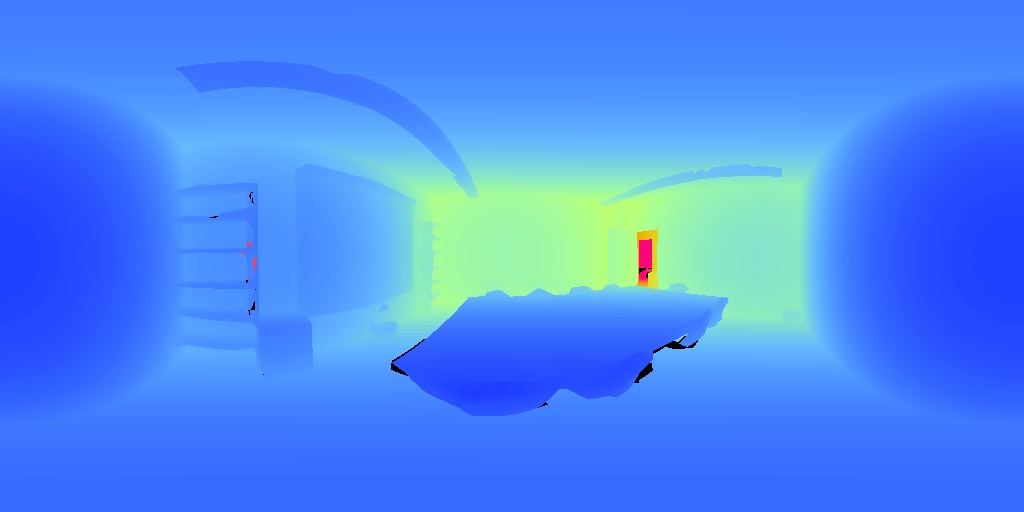} &
\includegraphics[height=\turnheightnew]{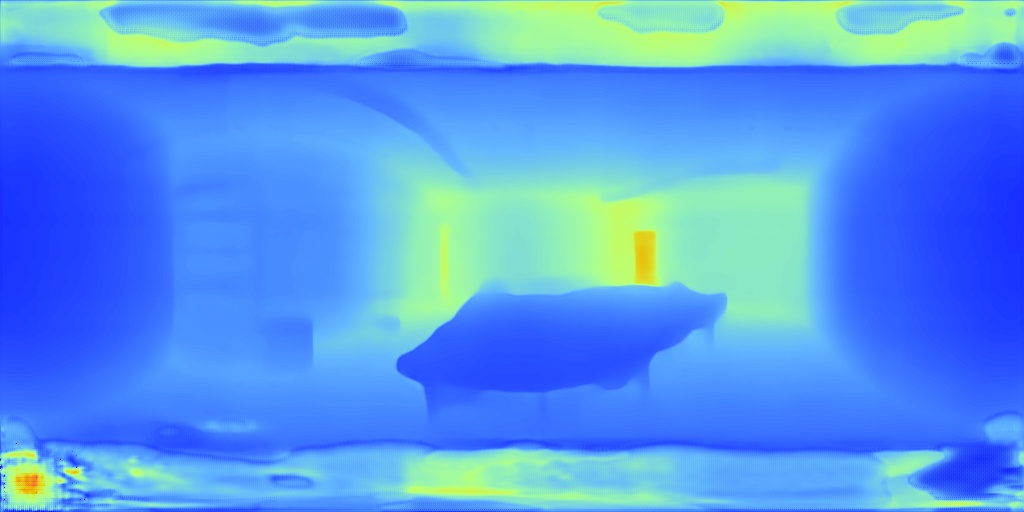} &
\includegraphics[height=\turnheightnew]{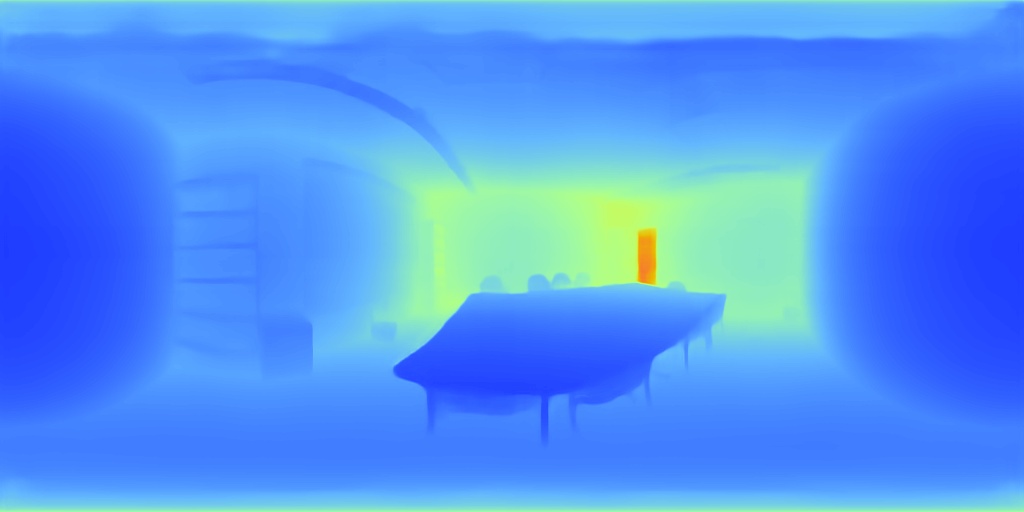}\\

{\rotatebox{90}{\hspace{0.1mm}\scriptsize Stanford2D3D}} &
\includegraphics[height=\turnheightnew]{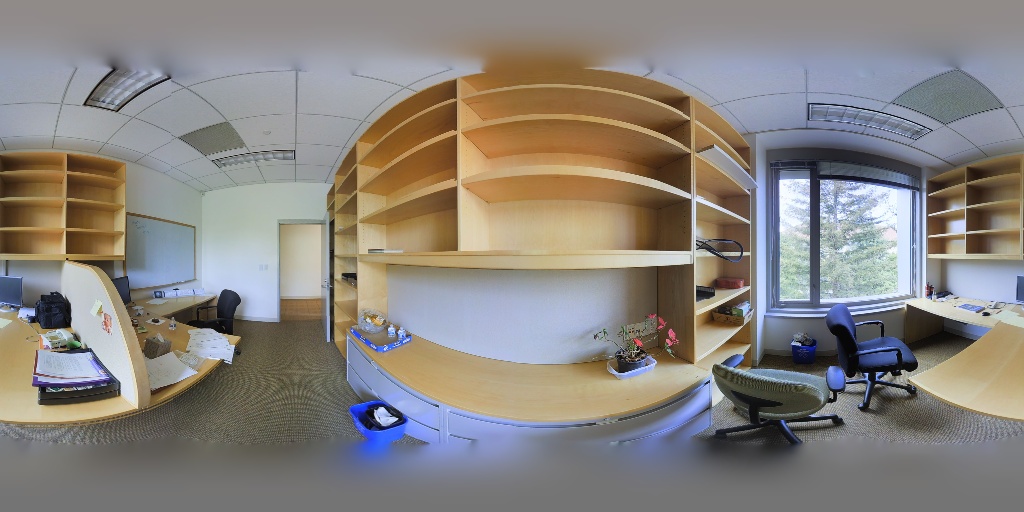} &
\includegraphics[height=\turnheightnew]{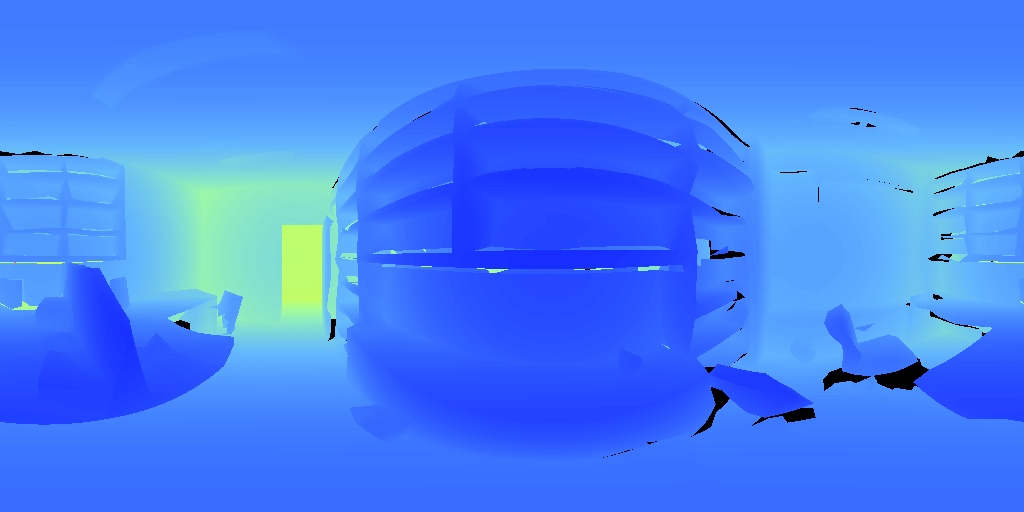} &
\includegraphics[height=\turnheightnew]{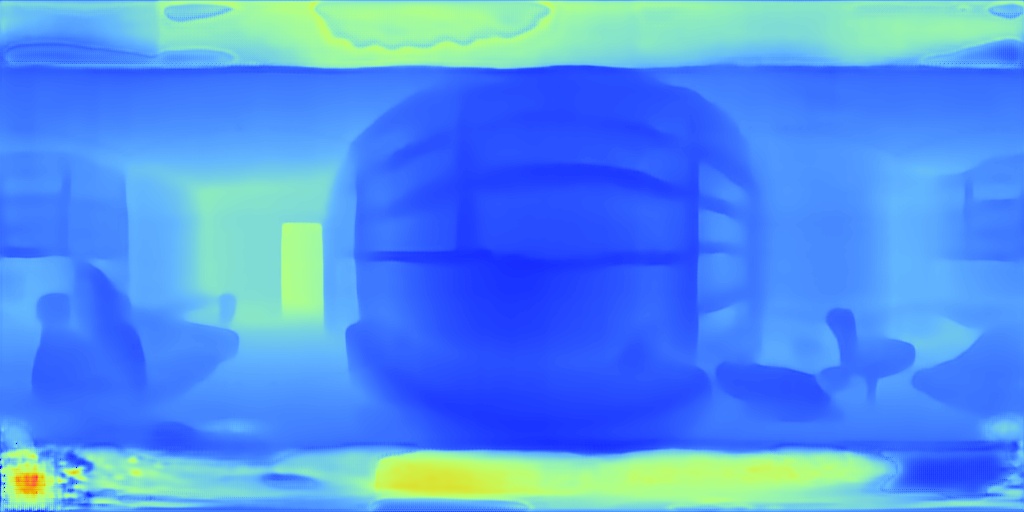} &
\includegraphics[height=\turnheightnew]{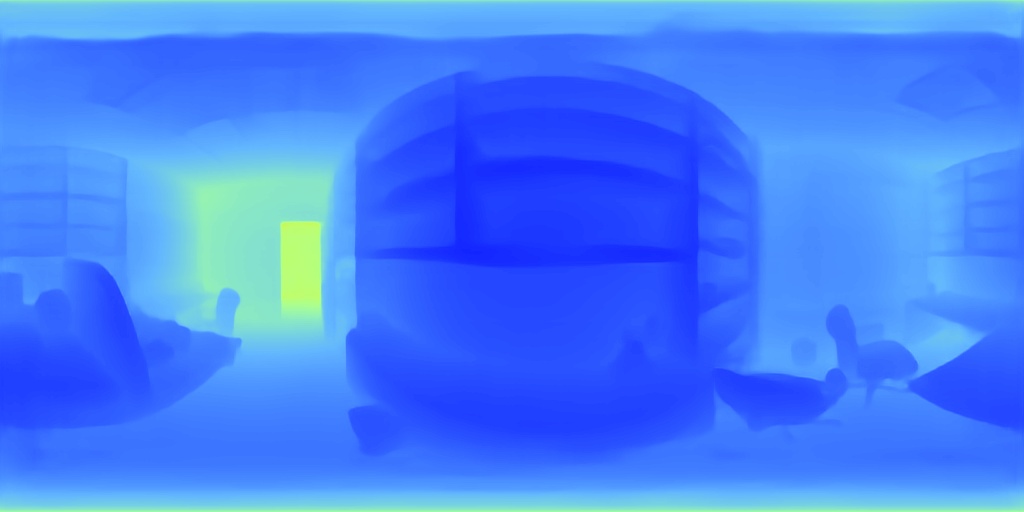}\\

{\rotatebox{90}{\hspace{0.1mm}\scriptsize Stanford2D3D}} &
\includegraphics[height=\turnheightnew]{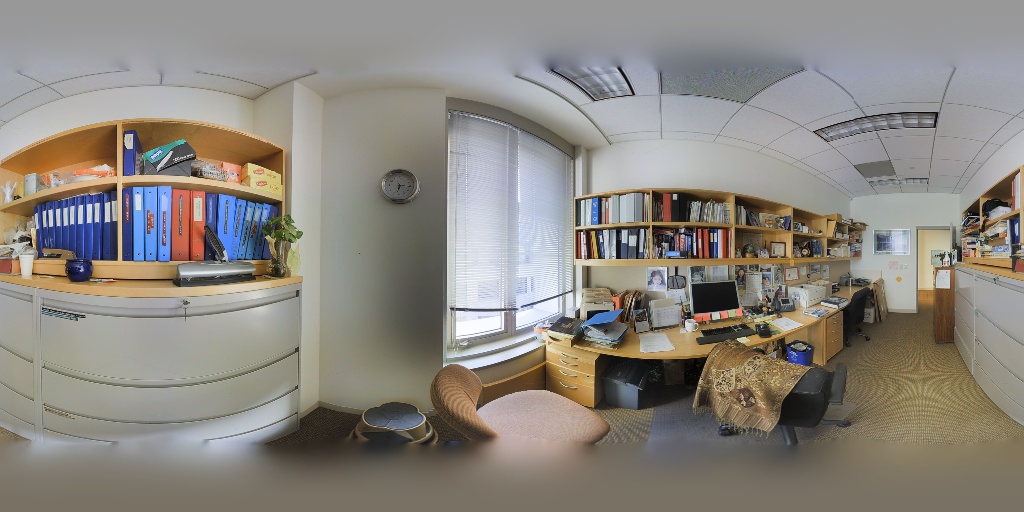} &
\includegraphics[height=\turnheightnew]{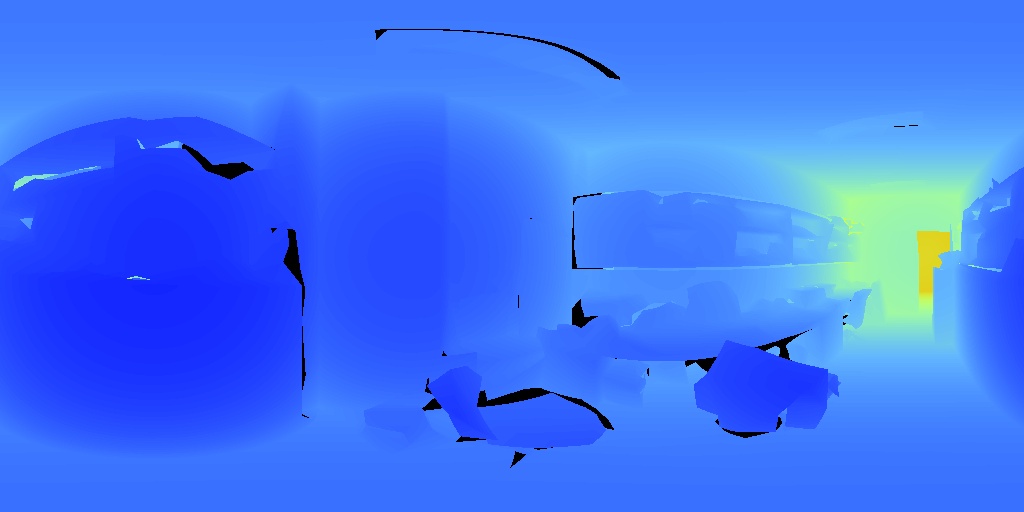} &
\includegraphics[height=\turnheightnew]{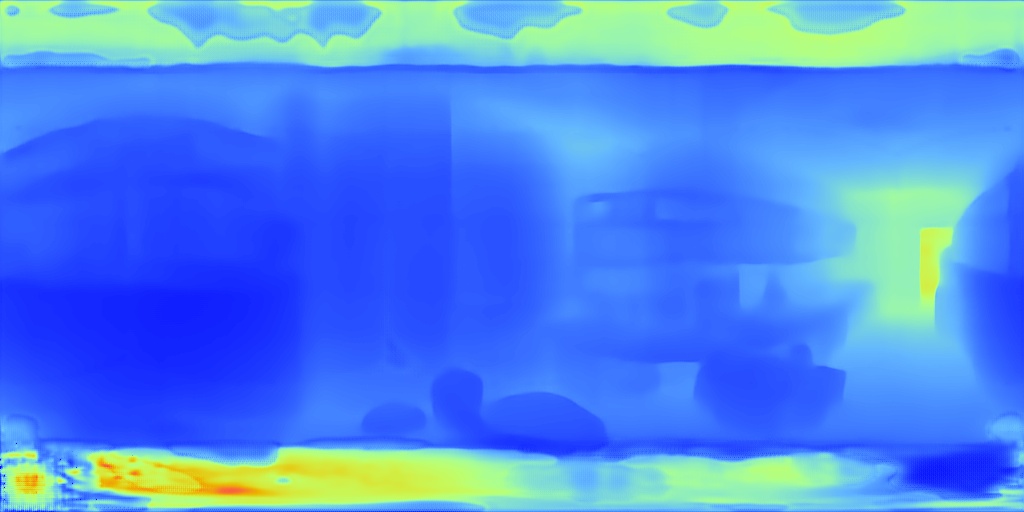} &
\includegraphics[height=\turnheightnew]{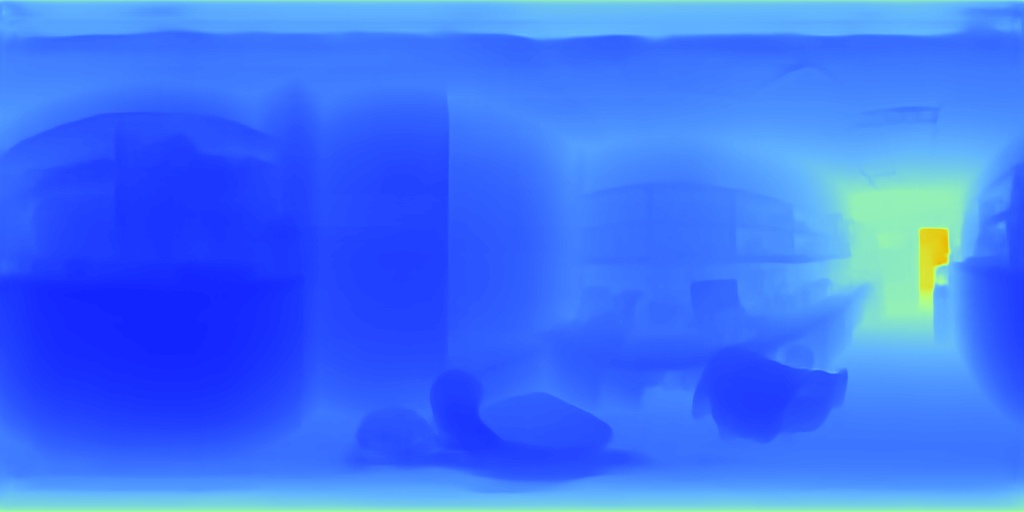}\\

&
\scriptsize Input &
\scriptsize Ground Truth&
\scriptsize BiFuse~\cite{wang2020bifuse} &
\scriptsize Our UniFuse\\

\end{tabular}
 }
\caption{\textbf{Qualitative Comparison between BiFuse and Our UniFuse Model.}  Best viewed in color.}
\label{fig:comparison2}
\end{figure*}

\subsubsection{Generalization Capability}
We examine the generalization capability between BiFuse~\cite{wang2020bifuse} and our UniFuse in Tab~\ref{tab:generalization}. 
We can perform such examination as BiFuse provides a pretrained model on the \textit{entire} Matterport3D dataset. We also trained our UniFuse on the whole Matterport3D dataset. Next, we evaluate both the BiFuse and our UniFuse on the test set of Matterport3D to see the effectiveness of fitting, as the test set has also been trained on. Finally, we evaluate both models on the test set of Stanford2D3D.  As the sensing depth of Matterport3D does not cover the top and bottom of the panorama, to make a fair comparison, we do not count the topmost and lowest 68 pixels following BiFuse's code when evaluating on Stanford2D3D. 
These two datasets have a certain domain gap, as Matterport2D3D is about various household scenes, while Stanford2D3D is about the office scenes of a university.
Thus, this transfer can examine the generalization capability of different models. 

From Tab~\ref{tab:generalization}, BiFuse's Abs Rel is 3 times of our UniFuse's and UniFuse has $8.0\%$ higher on the accuracy $\delta<1.25$  than BiFuse. This indicates UniFuse fits Matterport3D much better than BiFuse. But the better fitting is not overfitting, as UniFuse also transfers to Stanford2D3D much better. 
The visualization results on Fig~\ref{fig:comparison2} also verify this fact, and one extra merit of UniFuse is that even there is no ground truth depth on the top and bottom, it can produce plausible depth.

\begin{table}[t]
\vspace{5pt}
  \centering
  \resizebox{0.48\textwidth}{!}{
  \begin{tabular}{c | l |ccc}
  \toprule
  {Dataset} & {Method} & Abs Rel $\downarrow$ & RMSE $\downarrow$ & $\delta < 1.25 \uparrow $  \\
  \hline

  \multirow{2}{2cm}{\centering train on Matterport3D} &
  BiFuse~\cite{wang2020bifuse} & 0.1014  & 0.4070  & 90.48 \\
 
  & Our UniFuse & 0.0348 & 0.1863 & 98.47 \\
  \hline

  \multirow{2}{2cm}{\centering transfer to Stanford2D3D} &
   BiFuse~\cite{wang2020bifuse}& 0.1195 & 0.4339 & 86.16  \\
   & Our UniFuse  & 0.0944 & 0.3800 &  91.31 \\
  \bottomrule
  \end{tabular}
  }
\caption{\textbf{Generalization Comparison.}}
\label{tab:generalization}
\vspace{-10pt}
\end{table}

\section{CONCLUSIONS}
In this paper, we have shown our UniFuse model for single spherical panorama depth estimation. Our UniFuse model is a simple yet effective framework that utilizes both equirectangular projection and cubemap projection for $360^{\circ}$ depth estimation. We have also designed the new CEE fusion module for our framework to enhance the equirectangular features better. 
Experiments have verified that both our framework and module are effective. The final UniFuse model makes significant progress over the state-of-the-art methods on four popular $360^{\circ}$ panoramic datasets, especially on the biggest realistic dataset, Matterport3D. 
Furthermore, we have shown that our model has much lower model complexity and higher generalization capability than previous works, indicating the potential to apply it in real-world applications.
We are exploring the possibility to apply our model to practical fields, such as mobile robots.

{\small
\bibliographystyle{IEEEtran}
\bibliography{ref}
}

\end{document}